\documentclass[runningheads]{llncs}
\usepackage{graphicx}
\usepackage{enumerate}
\usepackage{multirow}
\usepackage{booktabs}

\usepackage[colorlinks,
            linkcolor=red,
            anchorcolor=blue,
            citecolor=green]{hyperref}

\usepackage{tikz}
\usepackage{comment}
\usepackage{amsmath,amssymb} 
\usepackage{color}
\usepackage{mathtools}
\usepackage{subcaption}

\DeclarePairedDelimiterX{\infdivx}[2]{(}{)}{%
  #1\;\delimsize\|\;#2%
}
\newcommand{\infdiv}{\mathrm {KL}\infdivx}

\definecolor{gain}{rgb}{0.224, 0.710, 0.290}
\newcommand{\gain}[1]{\textbf{\color{gain}{(+#1)}}}

\usepackage[misc,geometry]{ifsym} 

\usepackage[accsupp]{axessibility} 

\begin{document}

\pagestyle{headings}
\mainmatter
\def\ECCVSubNumber{6471} 

\title{FOSTER: Feature Boosting and  Compression for Class-Incremental Learning}

\author{Fu-Yun Wang\and
Da-Wei Zhou\and
Han-Jia Ye\textsuperscript{\Letter} \and
De-Chuan Zhan}

\institute{State Key Laboratory for Novel Software Technology, Nanjing University
\email{wangfuyun@smail.nju.edu.cn,\{zhoudw, yehj, zhandc\}@lamda.nju.edu.cn}}

\maketitle
\begin{abstract}
The ability to learn new concepts continually is necessary in this ever-changing world. However, deep neural networks suffer from catastrophic forgetting when learning new categories. Many works have been proposed to alleviate this phenomenon, whereas most of them either fall into the stability-plasticity dilemma or take too much computation or storage overhead.  Inspired by the gradient boosting algorithm to gradually fit the residuals between the target model and the previous ensemble model, we propose a novel two-stage learning paradigm FOSTER, empowering the model to learn new categories adaptively. Specifically, we first dynamically expand new modules to fit the residuals between the target and the output of the original model. Next, we remove redundant parameters and feature dimensions through an effective distillation strategy to maintain the single backbone model. We validate our method FOSTER on CIFAR-100 and ImageNet-100/1000 under different settings. Experimental results show that our method achieves state-of-the-art performance. Code is available at \url{https://github.com/G-U-N/ECCV22-FOSTER}.
\end{abstract}

\section{Introduction}\label{sec:intro}

The real world is constantly changing, with new concepts and categories continuously springing up~\cite{golab2003issues,zhou2021learning,wang2021afec,zhou2021learning2}. Retraining a model every time new classes emerge is impractical due to data privacy~\cite{delange2021continual} and expensive training costs. Therefore, it is necessary to enable the model to continuously learn new categories, namely class-incremental learning~\cite{zhou2022forward,zhou2022few,wang2021ordisco}. However, directly fine-tuning the original neural networks on new data causes a severe problem known as catastrophic forgetting~\cite{catastrophicforgetting} that the model entirely and abruptly forgets previously learned information. Inspired by this, class-incremental learning aims to design a learning paradigm that enables the model to continuously learn novel categories in multiple stages while maintaining the discrimination ability for old classes. 

In recent years, many approaches have been proposed from different aspects. So far, the most widely recognized and utilized class-incremental learning strategy is based on knowledge distillation~\cite{KD}. Methods~\cite{LWF,icarl,EEIL,bic,WA,zhou2021co} retain an old model additionally and use knowledge distillation to constrain output for original tasks of the new model to be similar to that of the old one~\cite{LWF}.    However, these methods with a single backbone may not have enough plasticity~\cite{stabilityplasticity} to cope with the coming new categories. Besides, even with restrictions of KD, the model still suffer from feature degradation~\cite{der} of old concepts due to limited access~\cite{delange2021continual}  to old data. Recently, methods~\cite{der,simpleder,dytox} based on dynamic architectures achieve state-of-the-art performance in class-incremental learning. Typically, they preserve some modules with their parameters frozen to maintain important sections for old categories and expand new trainable modules to strengthen plasticity for learning new categories. Nevertheless, they have two inevitable defects: First, constantly expanding new modules for coming tasks will lead to a drastic increase in the number of parameters, resulting in severe storage and computation overhead, which makes these methods not suitable for long-term incremental learning. Second, since old modules have never seen new concepts,  directly retaining them may harm performance in new categories. The more old modules kept, the more remarkable the negative impact.

\begin{figure}[t]
    \centering
    \includegraphics[width=12cm]{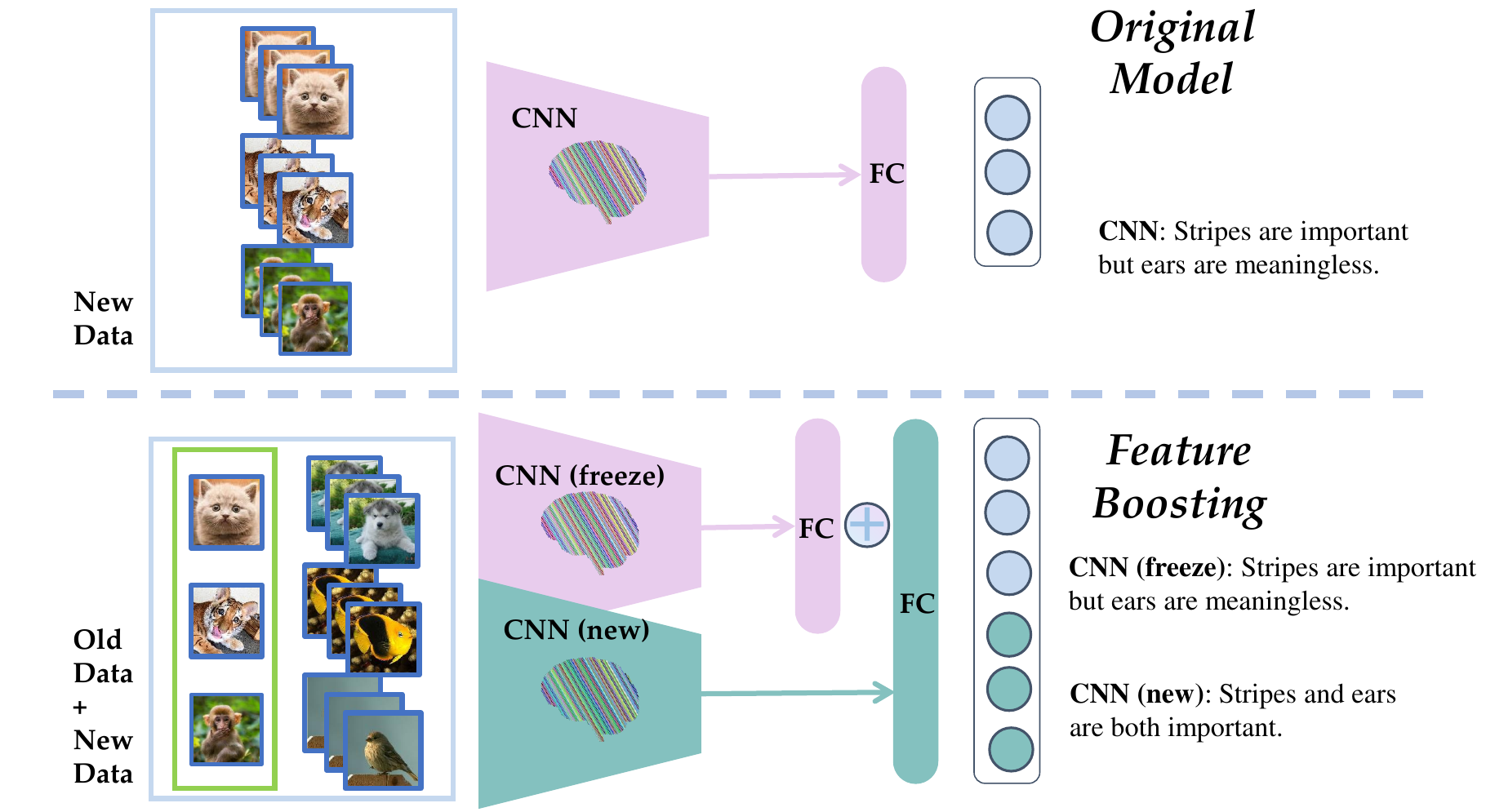}
    \caption{\small \textbf{Feature Boosting.} Illustration of feature boosting. When the task comes, we freeze the old model and create a new module to fit the residuals between the target and the output. The new module helps the model learn both new and old classes better. }
    \label{fig:boosting}
\end{figure}

 In this paper, we propose a novel perspective from gradient boosting to analyze and achieve the goal of class-incremental learning.  Gradient boosting methods use the additive model to gradually converge the ground-truth target model where the subsequent one fits the residuals between the target and the prior one. In class-incremental learning, since distributions of new categories are constantly coming, the distribution drift will also lead to the residuals between the target label and model output. Therefore, we propose a similar boosting framework to solve the problem of class-incremental learning by applying an additive model, gradually fitting residuals, where different models mainly handle their special tasks~(with nonoverlapping sets of classes). And as we discuss later, our boosting framework is a more generalized framework for dynamic structure methods~(\textit{e.g.}, DER\cite{der}). It has positive significance in two aspects: On the one hand, the new model enhances the plasticity and thus helps the model learn to distinguish between new classes. On the other hand, training the new model to classify all categories might contribute to discovering some critical elements ignored by the original model. As shown in Fig.~\ref{fig:boosting}, when the model learns old categories, including tigers, cats, and monkeys, it may think that stripes are essential information but mistakenly regard auricles as meaningless features. When learning new categories, because the fish and birds do not have auricles, the new model will discover this mistake and correct it. 

However, as we discussed above, creating new models not only leads to an increase in the number of parameters but also might cause inconsistency between the old and the new model at the feature level. To this end, we compress the boosting model to remove unnecessary parameters and inconsistent features, thus avoiding the above-mentioned drawbacks of dynamic structure-based methods, preserving crucial information, and enhancing the robustness of the model.  

In conclusion, our paradigm can be decoupled into two steps: boosting and compression. The first step can be seen as boosting to alleviate the performance decline due to the arrival of new classes. Specifically, we retain the old model with all its parameters frozen. Then we expand a trainable new feature extractor and concatenate it with the extractor of the old model and initialize a constrained, fully-connected layer to transform the super feature into logits, which we will demonstrate later in detail.  In the second step, we aim to eliminate redundant parameters and meaningless dimensions caused by feature boosting. Specifically, we propose an effective distillation strategy that can transfer knowledge from the boosting model to a single model with negligible performance loss, even if the data is limited when learning new tasks. Extensive experiments on three benchmarks, including CIFAR-100, ImageNet-100/1000 show that our method \textbf{F}eature Bo\textbf{O}\textbf{ST}ing and  Compr\textbf{E}ssion for class-inc\textbf{R}emental learning~(\textbf{FOSTER}) obtains the state-of-the-art performance.

\section{Related Work}

Many works have been done to analyze the reasons for performance degradation in class-incremental learning and alleviate this phenomenon. In this section, we will give a brief discussion of these methods and boosting algorithms.

\noindent\textbf{Knowledge Distillation.}
Knowledge distillation~\cite{KD} aims to transfer dark knowledge~\cite{korattikara2015bayesian} from the teacher to the student by encouraging the outputs of the student model to approximate the outputs of the teacher model~\cite{LWF}.  LwF~\cite{LWF} retains an old model additionally and applies a modified cross-entropy loss to constrain the outputs for old categories of the new model to preserve the capability for the old one. Bic~\cite{bic}, WA~\cite{WA} propose effective strategies to alleviate the bias of the classifier caused by imbalanced training data after distillation.

\noindent\textbf{Rehearsal.}
The rehearsal strategy enables the model to have partial access to old data. \cite{icarl,bic,WA,wang2021memory} allocate a memory to store exemplars of previous tasks for replay when learning tasks. \cite{iscen2020memory} preserves low dimensional features instead of raw instances to reduce the storage overhead. In~\cite{wu2021incremental}, instances are synthesized by a generative model~\cite{GANs} for rehearsal. \cite{CILsurvey} test various exemplar selection strategies, showing that different ways of exemplar selection have a significant impact on performance and herding surpass other strategies in most settings. 

\noindent\textbf{Dynamic Architectures.}
Many works~\cite{fernando2017pathnet,golkar2019continual,hung2019compacting,rusu2016progressive,wen2020batchensemble} create new modules to handle the growing training distribution~\cite{yoon2017lifelong,lesort2019generative} dynamically. However, an accurate task id, which is usually unavailable in real-life, is needed for most of these approaches to help them choose the corresponding id-specific module. Recently, methods~\cite{der,simpleder,dytox} successfully apply the dynamic architectures into class incremental learning where the task id is unavailable, showing their advantages over the single backbone methods.  However, as we illustrate in Sec.~\ref{sec:intro}, they have two unavoidable shortcomings: $(\textrm{i})$ Continually adding new modules causes unaffordable overhead. $(\textrm{ii})$ Directly retaining old modules leads to noise in the representations of new categories, harming the performance in new classes. 

\noindent\textbf{Boosting.}
Boosting represents a family of machine learning algorithms that convert weak learners to strong ones~\cite{zhou2012ensemble}. AdaBoost~\cite{friedman2000additive} is one of the most famous boosting algorithms, aiming to minimize the exponential loss of the additive model. The crucial idea of AdaBoost is to adjust the weights of training samples to make the new base learner pay more attention to samples that the former ensemble model cannot recognize correctly. In recent years, gradient boosting~\cite{friedman2001greedy} based algorithms ~\cite{chen2016xgboost,ke2017lightgbm,dorogush2018catboost} achieve excellent performance on various tasks.

\section{Preliminary}
In this section, we first briefly discuss the basic process of gradient boosting in Sec.~\ref{sec:GB}. Then, we describe the setting of class-incremental learning in Sec.~\ref{sec:setup}.  In Sec.~\ref{sec:method}, we will give an explicit demonstration of how we apply the idea of gradient boosting to the scenario of class-incremental learning.

\subsection{Gradient Boosting}\label{sec:GB}

Given a training set $\mathcal D_{train}=\{(x_i,y_i)\}_{i=1}^n$, where $x_i \in \mathcal X$ is the instance and $y_i \in \mathcal Y$ is the corresponding label, the gradient boosting methods seek a hypothesis $\mathrm F: \mathcal X \rightarrow \mathcal Y$ to minimize the empirical risk~(with loss function $\ell (\cdot, \cdot)$)
\begin{equation}
\mathrm F^*= \mathop{\arg\min}_\mathrm F \mathbb E_{(x,y)\in \mathcal D_{train}}\left[\ell\left(y,\mathrm F(x)\right)\right]\, ,
\end{equation}
by iteratively adding a new weighted weak function $h_i(\cdot)$ chosen from a specific function space $\mathcal H_i$~(\textit{e.g}., the set of all possible decision trees) to gradually fit residuals. After $m$ iterations, the hypothesis $\mathrm F$ can be represented as 
\begin{equation}
\mathrm F(x) = \mathrm F_m(x)= \sum_{i=1}^m\alpha_ih_i(x)\, ,
 \end{equation}
where $\alpha_i$ is the coefficient of $h_i(\cdot)$. Then we are supposed to find $\mathrm F_{m+1}$ for further optimization of the objective \begin{equation}
\mathrm F_{m+1}(x)= \mathrm F_m(x)+\mathop{\arg\min}_{h_{m+1}\in\mathcal H_{m+1}} \mathbb E_{(x,y)\in \mathcal D_{train}}\left[\ell\left(y,\mathrm F_{m}(x)+h_{m+1}(x)\right)\right]\, .
\end{equation}
However, directly optimizing the above function to find the best $h_{m+1}$ is typically infeasible. Therefore, we use the steepest descent step for iterative optimization:
\begin{equation}
\mathrm F_{m+1}(x)=\mathrm F_{m}(x)-\alpha_{m}\nabla_{\mathrm F_m}\mathbb E_{(x,y)\in \mathcal D_{train}}\left[\ell\left(y,\mathrm F_{m}(x)\right)\right]\, ,
\end{equation}
where $-\nabla_{\mathrm F_m}\mathbb E_{(x,y)\in \mathcal D_{train}}\left[\ell\left(y,\mathrm F_{m}(x)\right)\right]$ is the objective for $h_{m+1}(x)$ to approximate. Specifically, if $\ell(\cdot,\cdot)$ is the mean-squared error~(MSE), it transforms into
\begin{equation}
-\nabla_{\mathrm F_m}\mathbb E_{(x,y)\in \mathcal D_{train}}\left[\left(y-\mathrm F_{m}(x)\right)^2\right]=2\times \mathbb E_{(x,y)\in \mathcal D_{train}}\left[y-\mathrm F_{m}(x)\right]\, .
\end{equation}
Ideally, let $\alpha_m=1/2$, if $h_{m+1}(x)$ can fit $2\alpha _m(y-\mathrm F_{m}(x))=(y-\mathrm F_m(x))$ for each $(x,y)\in \mathcal D_{train}$,  $\mathrm F_{m+1}$ is the optimal function, minimizing the empirical error.  
\subsection{Class-Incremental Learning Setup }\label{sec:setup}
Unlike the traditional case where the model is trained on all classes with all training data available, in class-incremental learning, the model receives a batch of new training data $\mathcal{D}_t=\{(\boldsymbol x_i^t, y_i^t)\}_{i=1}^{n}$ in the $t^\text{th}$ stage. Specifically, $n$ is the number of training samples, $\boldsymbol x_i^t \in \mathcal X_t$ is the input image, and $y_i^t \in \mathcal Y_t$ is the corresponding label for $\boldsymbol x_i^t$. Label space of all seen categories is denoted as $\hat{\mathcal Y}_t=\cup_{i=0}^t\mathcal Y_i$, where $\mathcal Y_t \cap \mathcal Y_{t^\prime}=\emptyset$ for $t\not=t^\prime$. In the $t^\text{th}$ stage, rehearsal-based methods also save a part of old data as $\mathcal V_{t}$, a limited subset of $\cup_{i=0}^{t-1} \mathcal D_{i}$. Our model is trained on $\hat{\mathcal D}_t=\mathcal D_t \cup \mathcal V_t$ and is required to perform well on all seen categories. 

\section{Method}\label{sec:method}

In this section, we give a description of FOSTER and how it works to prompt the model to simultaneously learn all classes well. Below, we first give a full demonstration of how the idea of the gradient boosting algorithm is applied to class-incremental learning in Sec.~\ref{sec:GB2CIL}.  Then we propose novel strategies to further enhance and balance  the learning, which greatly improves the performance in Sec.~\ref{sec:calibration}. Finally, in order to avoid the explosive growth of parameters and remove redundant parameters and feature dimensions, we utilize a straightforward and effective compression method based on knowledge distillation in Sec.~\ref{sec:compression}.

\subsection{From Gradient Boosting to Class-Incremental Learning}\label{sec:GB2CIL}
Assuming in the $t^\text{th}$ stage, we have saved the model $\mathrm F_{t-1}$ from the last stage. $\mathrm F_{t-1}$ can be further decomposed into feature embedding and linear classifier: $\mathrm F_{t-1}(\boldsymbol x)=(\mathbf W_{t-1})^\top\mathrm \Phi_{t-1}(\boldsymbol x)$, where $\mathrm \Phi_{t-1}(\cdot):\mathbb R^D\rightarrow \mathbb R^d$ and $\mathbf W_{t-1}\in \mathbb R^{d\times \lvert\hat{\mathcal Y}_{t-1}\rvert}$. When a new data stream comes,  directly fine-tuning $\mathrm F_{t-1}$ on the new data will impair its capacity for old classes, which is inadvisable. On the other hand, simply freezing $\mathrm F_{t-1}$ causes it to lose plasticity for new classes, making the residuals between target $y$ and $\mathrm F_{t-1}(\boldsymbol x)$ large for $(\boldsymbol x,y) \in \mathcal D_{t}$. Inspired by gradient boosting, we train a new model to fit the residuals. Specifically, the new model $\mathcal F_{t}$ consists of a feature extractor $ \phi_{t}(\cdot):\mathbb R^D\rightarrow \mathbb R^d$ and a linear classifier $\mathcal W_t\in \mathbb R^{d\times \lvert \hat{\mathcal {Y}}_{t}\rvert}$. $\mathcal W_{t}$ can be further decomposed into $\left[\mathcal W_t^{(o)},\mathcal W_{t}^{(n)}\right]$, where $\mathcal W_{t}^{(o)}\in \mathbb R^{d\times \lvert\hat{\mathcal Y}_{t-1}\rvert }$ and $\mathcal W_{t}^{(n)}\in \mathbb R^{d\times \lvert \mathcal Y_t\rvert}$ . Accordingly, the training process can be represented as 
\begin{equation}
\mathrm F_{t}(\boldsymbol x) = \mathrm F_{t-1}(\boldsymbol x)+\mathop{\arg\min}_{\mathcal F_{t}} \mathbb E_{(\boldsymbol x,y)\in \hat{\mathcal D}_{t}}\left[\ell\left(y,\mathrm F_{t-1}(\boldsymbol x)+\mathcal F_{t}(\boldsymbol x)\right)\right]\, .
\end{equation}
Similar to Sec.~\ref{sec:GB}, let $\ell(\cdot,\cdot)$ be the mean-squared error function, considering the strong feature representation learning ability of neural networks, we expect $\mathcal F_t(\boldsymbol x)$ can fit residuals of $ y$ and $\mathrm F_{t-1}(\boldsymbol x)$ for every $(\boldsymbol x, y)\in \hat{\mathcal D}_{t}$. Ideally, we have
\begin{equation}
    \boldsymbol y=\mathrm F_{t-1}(\boldsymbol x)+\mathcal F_{t}(\boldsymbol x)=\mathcal S\left(\left[\begin{array}{cc}\mathbf W_{t-1}^\top\\ \mathbf O\end{array}\right] \mathrm \Phi_{t-1}(\boldsymbol x)\right)+\mathcal S\left(\left[\begin{array}{cc}(\mathcal W^{(o)}_{t})^\top\\ (\mathcal W^{(n)}_{t})^\top\end{array}\right] \phi_{t}(\boldsymbol x)\right)\, ,
\end{equation}
where $\mathcal S(\cdot)$ is the softmax operation, $\mathbf O\in \mathbb R^{d\times \lvert\mathcal Y_t\rvert}$ is set to zero matrix or fine-tuned on $\hat{\mathcal D}_{t}$ with $\Phi_{t-1}$ frozen, and $\boldsymbol y$ is the corresponding one-hot vector of $y$. We set $\mathbf O$ to  zero  matrix as default in our discussion.

Denote the parameters of  $\mathcal F_{t}$ as $\theta_t$ and $\textrm{Dis}(\cdot , \cdot)$ as a distance metric~(\textit{e.g}., euclidean metric),  this process can be represented as the following optimization problem:
\begin{equation}
\theta_t^*=\mathop{\arg\min}_{\theta_t}  \textrm{Dis}\left(\boldsymbol y, \mathcal S\left(\left[\begin{array}{cc}\mathbf W_{t-1}^\top\\ \mathbf O\end{array}\right] \mathrm \Phi_{t-1}(\boldsymbol x)\right)+\mathcal S\left(\left[\begin{array}{cc}(\mathcal W^{(o)}_{t})^\top\\ (\mathcal W^{(n)}_{t})^\top\end{array}\right]\mathcal \phi_{t}(\boldsymbol x)\right)\right)\, .
\end{equation}
We replace the $\mathcal S(\cdot) + \mathcal S(\cdot)$ with $\mathcal S(\cdot+\cdot)$ and substitute the $\textrm{Dis}(\cdot,\cdot)$ for the Kullback-Leibler divergence~(KLD), then the objective function changes into:
\begin{equation}
    \theta_{t}^*=\mathop{\arg\min}_{\theta_t}\infdiv*{y}{ \mathcal S\left(\left[\begin{array}{cc}\mathbf W_{t-1}^\top & (\mathcal W_{t}^{(o)})^\top\\
\mathbf O& (\mathcal W_{t}^{(n)})^\top\end{array}\right]\left[\begin{array}{cc}\mathrm \Phi_{t-1}(\boldsymbol x)\\\mathcal \phi_{t}(\boldsymbol x)\end{array}\right]\right)}\, .
\end{equation}
We provide an illustration about the reasons for this substitution in the supplementary material. Therefore, $\mathrm F_{t}$  can be further decomposed as an expanded linear classifier $\mathbf W_t$ and a concatenated super feature extractor $\mathrm \Phi_{t}(\cdot)$, where
\begin{align}
\mathbf W_t^\top&= \left[\begin{array}{cc}\mathbf W_{t-1}^\top & (\mathcal W_{t}^{(o)})^\top\\
\mathbf O& (\mathcal W_{t}^{(n)})^\top\end{array}\right]\, , &
\mathrm \Phi_{t}(\boldsymbol x)&=\left[\begin{array}{cc}\mathrm \Phi_{t-1}(\boldsymbol x)\\\mathcal \phi_{t}(\boldsymbol x)\end{array}\right]\, .
\end{align}
Note that $\mathbf W_{t-1}^\top$, $\mathbf O$, and $\mathrm \Phi_{t-1}$ are all frozen, the trainable modules are the $\mathcal \phi_{t}, \mathcal W_{t}^{(o)}, \mathcal W_{t}^{(n)}$. Here we explain their roles. Eventually, logits of $\mathrm F_t$ is 
\begin{align}\label{equation:logits}
\mathbf W_{t}^\top \mathrm \Phi_{t}(\boldsymbol x)=\left[\begin{array}{cc}\mathbf W_{t-1}^\top \mathrm \Phi_{t-1}(\boldsymbol x)+(\mathcal W_{t}^{(o)})^\top\mathcal \phi_{t}(\boldsymbol x)\\
(\mathcal W_{t}^{(n)})^\top\mathcal \phi_{t}(\boldsymbol x)\end{array}\right].
\end{align}
The lower part is the logits of new classes, and the upper part is that of old ones. As we claimed in Sec.~\ref{sec:intro}, the lower part requires the new module  $\mathcal F_{t}$ to learn how to correctly classify new classes, thus enhancing the model's plasticity to redeem the performance on new classes. The upper part encourages the new module to fit the residuals between $y$ and $\mathrm F_{t-1}$, thus encouraging $\mathcal F_{t}$ to exploit more pivotal patterns for classification.

\subsection{Calibration for Old and New}\label{sec:calibration}
When training on new tasks, we only have an imbalanced training set $\hat{\mathcal D_{t}}=\mathcal D_{t}\cup \mathcal V_{t}$. The imbalance on categories of $\mathcal D_{t}$ will result in a strong classification bias in the model~\cite{Decouple,WA,bic,EEIL}. Besides, the boosting model tends to ignore the residuals of minor classes due to insufficient supervision. To alleviate the classification bias and encourage the model to equally learn old and new classes, we propose Logits Alignment and Feature Enhancement strategies in the following sections.

\noindent\textbf{Logits Alignment.}\label{sec:LA}
To strengthen the learning of old instances and mitigate the classification bias, we add a scale factor to the logits of the old and new classes in Eq.~\ref{equation:logits} respectively during training.  Thus, the logits during training are:
\begin{equation}
\boldsymbol \gamma\mathbf W_{t}^\top \mathrm \Phi_{t}(\boldsymbol x)=\left[\begin{array}{cc}\gamma_1\left(\mathbf W_{t-1}^\top \mathrm \Phi_{t-1}(\boldsymbol x)+(\mathcal W_{t}^{(o)})^\top\mathcal \phi_{t}(\boldsymbol x)\right)\\
\gamma_2(\mathcal W_{t}^{(n)})^\top\mathcal \phi_{t}(\boldsymbol x)\end{array}\right]\, ,
\end{equation}
where $0<\gamma_1<1$, $\gamma_2>1$, and $\boldsymbol \gamma$ is a diagonal matrix composed of $\gamma_1$ and $\gamma_2$. Through this scaling strategy, the absolute value of logits for old categories is reduced, and the absolute value of logits for new ones is enlarged, thus forcing the model $\mathrm F_{t}$ to produce larger logits for old categories and smaller logits for new categories. 

We get the scale factors $\gamma_1, \gamma_2$ trough the normalized effective number $E_n$~\cite{CBCE} of each class, which can be seen as the summation of proportional series, where $n$ equal to the number of instances and $\beta$ is an adjustable hyperparameter
\begin{equation}\label{equation:effectivenumber}
    E_n = \begin{cases}
    \frac{1-\beta^n}{1-\beta},&\beta\in [0,1)\\
    n, & \beta=1
    \end{cases}\, ,
\end{equation}
concretely, $\left(\gamma_1, \gamma_2\right)=\left(\frac{E_{n_\text{old}}}{E_{n_\text{old}}+E_{n_\text{new}}}, \frac{E_{n_\text{new}}}{E_{n_\text{old}}+E_{n_\text{new}}}\right)$.
Hence the objective is formulated as:
\begin{equation}\label{eq:la}
\mathcal L_{LA}=\infdiv*{y}{\mathcal S\left(\boldsymbol \gamma\mathbf W_{t}^\top \mathrm \Phi_{t}(\boldsymbol x)\right)}\, .    
\end{equation}

\noindent\textbf{Feature Enhancement.}\label{sec:FE}
We argue that simply letting a new module $\mathcal F_{t}(\boldsymbol x)$ fit the residuals of $\mathrm F_{t-1}(\boldsymbol x)$ and label $y$ is sometimes insufficient.  At the extreme,, for instance, the residuals of $\mathrm F_{t-1}(\boldsymbol x)$ and $y$ is zero. In that case, the new module $\mathcal F_{t}$ can not learn anything about old categories, and thus it will damage the performance of our model for old classes. Hence, we should prompt the new module $\mathcal F_{t}$ to learn old categories further.

Our Feature Enhancement consists of two parts. First, we initialize a new linear classifier $\mathbf W_{t}^{(a)}\in \mathbb R^{d\times \lvert \hat{\mathcal Y_{t}}\rvert}$ to transform the new feature $\phi_{t}(\boldsymbol x)$ into logits of all seen categories and require the new feature itself to correctly classify all of them:
\begin{equation}
\mathcal L_{FE}=\infdiv*{y}{\mathcal S\left((\mathbf W_{t}^{(a)})^\top \mathcal \phi_{t}(\boldsymbol x)\right)}\, .
\end{equation}
Hence, even if the residuals of $\mathrm F_{t-1}(\boldsymbol x)$ and $y$ is zero, the new feature extractor $\phi_{t}$ can still learn how to classify the old categories. Besides, it should be noted that simply using one-hot targets to train the new feature extractor in an imbalanced dataset might lead to overfitting to small classes, failing to learn a  feature representation with good generalization ability for old categories. To alleviate this phenomenon and provide more supervision for old classes, we utilize knowledge distillation to encourage $\mathrm F_{t}(\boldsymbol x)$ to have similar output distribution as $\mathrm F_{t-1}$ on old categories,
\begin{equation}
\mathcal L_{KD}=\infdiv*{\mathcal S\left(\mathrm F_{t-1}(\boldsymbol x)\right)}{\mathcal S\left(\mathrm F_{t-1}(\boldsymbol x)+(\mathcal W_{t}^{(o)})^\top\mathcal \phi_{t}(\boldsymbol x)\right)}\, .
\end{equation}
Note that this process requires only one more time matrix multiplication computation because the forward process of the original model $\mathrm F_{t-1}$ and the expanded model $\mathrm F_{t}$ are shared, except for the final linear classifier.

\noindent\textbf{Summary of Feature Boosting.}
To conclude, feature-boosting consists of three components. First, we create a new module to fit the residuals between targets and the output of the original model, following the principle of gradient boosting. With reasonable simplification and deduction,  the optimization objective is transformed into the minimization of KL divergence of the target and the output of the concatenated model. To alleviate the classification bias caused by imbalanced training, we proposed logits alignment~(LA) to balance the training of old and new classes. Moreover, we argued that simply letting the new module fit the residuals is sometimes insufficient. To further encourage the new module to learn old instances, we proposed feature enhancement, where $\mathcal L_{FE}$ aims to make the new module learn the difference among all categories by optimizing the cross-entropy loss of target and the output of the new module, and $\mathcal L_{KD}$ utilize the original output to instruct the expanded model through knowledge distillation.  
The final FOSTER loss for boosting combines the above three components:
\begin{equation}
    \mathcal L_{Boosting}=\mathcal L_{LA}+\mathcal L_{FE} +\mathcal L_{KD}\, . 
\end{equation}
\begin{figure}[t]
    \centering
    \includegraphics[width=11cm]{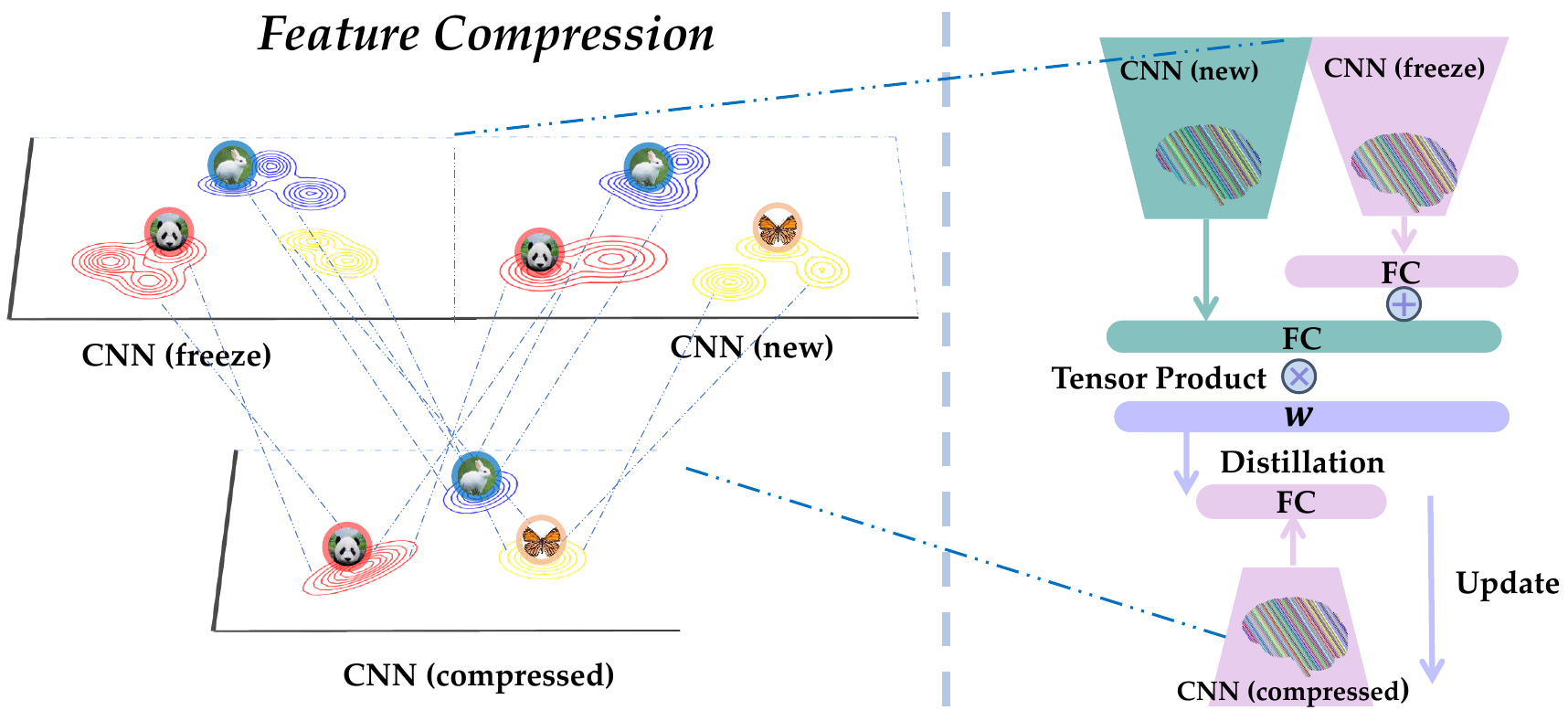}
    \caption{\small \textbf{Feature Compression.} Left: the process of feature compression. We remove insignificant dimensions and parameters to make the distribution of the same categories more compact. Right: the implementation of feature compression. Outputs of the dual branch model are used to instruct the representation learning of the compressed model. Different weights are assigned to old and new classes to alleviate the classification bias. }
    \label{fig:compression}
\end{figure}

\subsection{Feature Compression}\label{sec:compression}
Our method FOSTER achieves excellent performance through gradient boosting. However, gradually adding a new module $\mathcal F$ to our model $\mathrm F_t$ will lead to the growing number of parameters and feature dimensions of our model $\mathrm F_t$, making it unable to be applied in long-term incremental learning tasks. Do we really require so many parameters and feature dimensions? For example, we create the same module $\mathcal F$ to learn tasks with 2 classes and 50 classes and achieve similar effects. Thus, there must be redundant parameters and meaningless feature dimensions in the task with 2 classes. Are we able to compress the expanded feature space of $\mathrm F_t$ to a smaller one with almost no performance degradation? 
  
Knowledge distillation~\cite{KD} is a simple yet effective way to achieve this goal. Since our model $\mathrm F_t$ can handle all seen categories with excellent performance, it can give any input a soft target, namely the output distribution on all known categories. Therefore, except for the current training set $\hat{\mathcal D}_t$, we can sample other unlabeled data from a similar domain for further distillation. Note that these unlabeled data can be obtained from the Internet during distillation and discarded after that, so it does not occupy additional memory. 

Here, we do not expect any additional auxiliary data to be available and achieve remarkable performance with only the imbalanced dataset $\hat{\mathcal D}_t$.

\noindent\textbf{Balanced Distillation.}
Suppose there is a single backbone student model ${\mathrm F}^{(s)}_t$to be distilled.  To mitigate the classification bias caused by imbalanced training datasets $\hat{\mathcal D_t}$,  we should consider the class priors and adjust the weights of distilled information for different classes~\cite{zhang2021balanced}. Therefore, the Balanced Distillation loss is formulated as:
\begin{equation}\label{eq:bkd}
    \mathcal L_\text{BKD}=\infdiv*{\boldsymbol w\otimes\mathcal S\left(\mathrm F_{t}(\boldsymbol x)\right)}{\mathcal S(\mathrm F_{t}^{(s)}(\boldsymbol x))}\, ,
\end{equation}
where $\otimes$ means the tensor product~(\textit{i.e}., automatically broadcasting to different batchsizes.) and $\boldsymbol w$ is the weighted vector obtained from Eq.~\ref{equation:effectivenumber} to make classes with fewer instances have larger weights.

\section{Experiments}
In this section, we compare our  FOSTER with other SOTA methods on benchmark incremental learning datasets.  We also perform ablations to validate the effectiveness of FOSTER components and their robustness to hyperparameters.

\subsection{Experimental Settings}

\noindent\textbf{Datasets.}
We validate our methods on widely used benchmark of class-incremental learning CIFAR-100~\cite{cifar100} and ImageNet100/1000~\cite{imagenet1000}.  \textbf{CIFAR-100}: CIFAR-100 consists of 50,000 training images with 500 images per class, and 10,000 testing images with 100 images per class. \textbf{ImageNet-1000}: ImageNet-1000 is a large scale dataset composed of about 1.28 million images for training and 50,000 for validation with 500 images per class. \textbf{ImageNet-100}: ImageNet-100 is composed of 100 classes randomly chosen from the original ImageNet-1000.

\noindent\textbf{Protocol.}
For both the CIFAR-100 and ImageNet-100, we validate our method on two widely used protocols: \textrm{(i)} \textbf{CIFAR-100/ImageNet-100 B0~(base 0)}: In the first protocols, we train all 100 classes gradually with 5, 10, 20 classes per step with the fixed memory size of 2,000 exemplars. (\textrm{ii}) \textbf{CIFAR-100/ImageNet-100 B50~(base 50)}: We also start by training the models on half the classes. Then we train the rest 50 classes with 2, 5, 10 classes per step with 20 exemplars per class. For ImageNet-1000, we train all 1000 classes with 100 classes per step~(10 steps in total) with a fixed memory size of 20,000 exemplars.
\begin{figure*}[t]
  \centering
  \begin{subfigure}[b]{0.32\linewidth}
    \includegraphics[width=\linewidth]{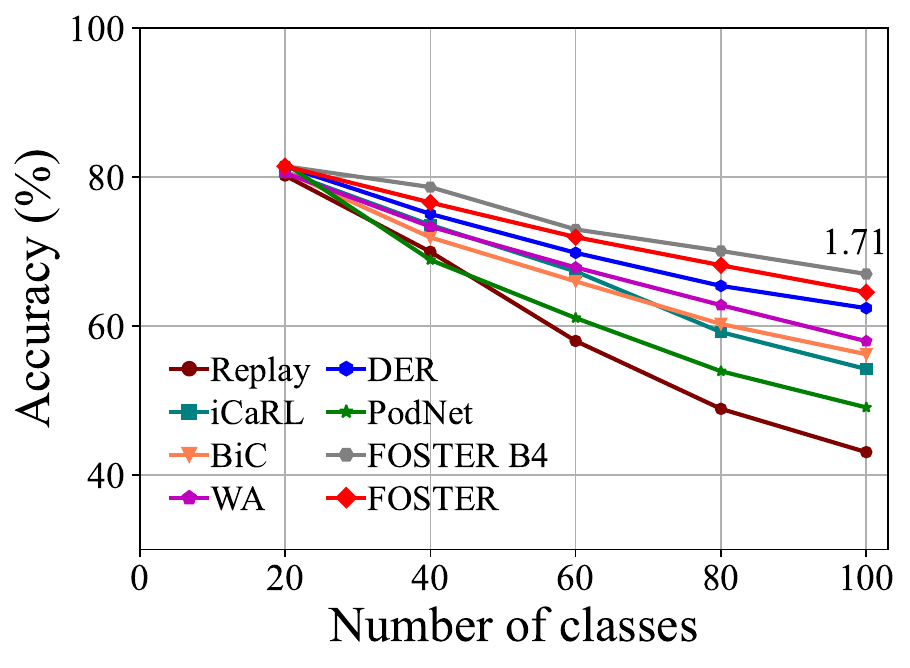}
    \caption{\scriptsize CIFAR-100 B0 5 steps}
    \label{fig:cifar_inc1}
  \end{subfigure}
  \hfill
  \begin{subfigure}[b]{0.32\linewidth}
    \includegraphics[width=\linewidth]{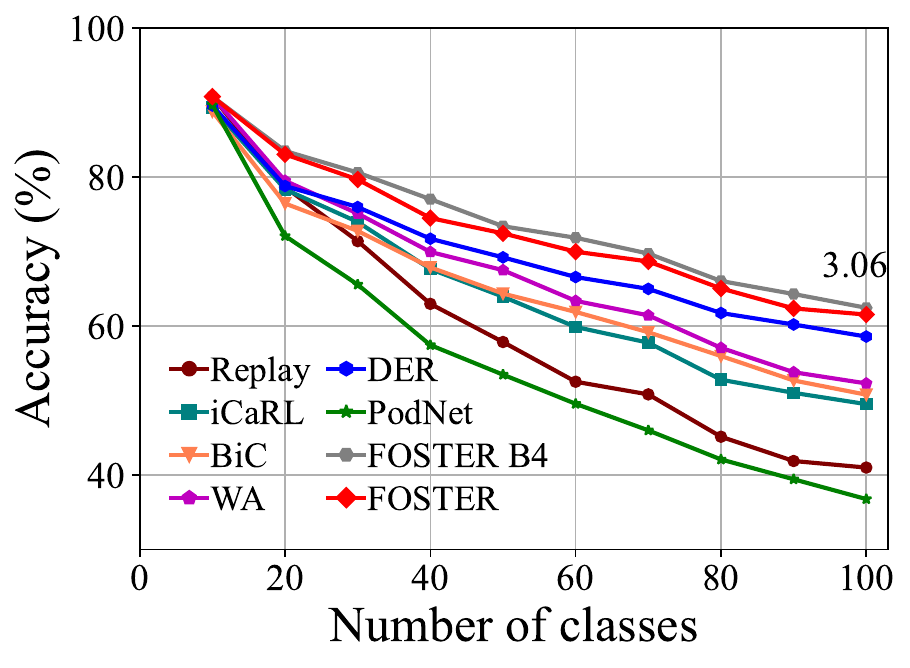}
    \caption{\scriptsize CIFAR-100 B0 10 steps}
    \label{fig:cifar_inc2}
  \end{subfigure}
   \hfill
  \begin{subfigure}[b]{0.32\linewidth}
    \includegraphics[width=\linewidth]{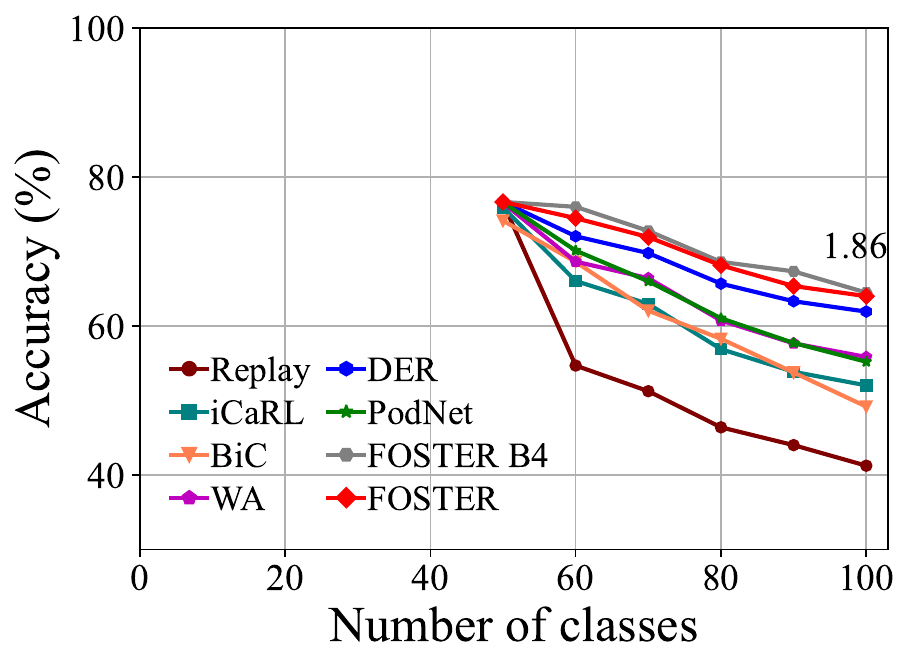}
    \caption{\scriptsize CIFAR-100 B50 5 steps}
    \label{fig:cifar_inc3}
  \end{subfigure}
  \caption{\small  \textbf{Incremental Accuracy on CIFAR-100.} Replay is the baseline with naive rehearsal strategy. FOSTER B4 records the accuracy of the dual branch model after feature boosting. FOSTER records the accuracy of the single backbone model after feature compression. The performance gap is annotated at the end of each curve.}
  \label{fig:cifar100}
\end{figure*}
\begin{table*}[t]
\caption{\small  Average incremental accuracy on CIFAR-100 for FOSTER vs. state-of-the-art. DER uses the same number of backbone models as incremental sessions, while the other methods, including FOSTER, retain only one backbone after each session.}
\label{tab:cifar100}
\centering
\setlength{\tabcolsep}{2mm}
\begin{tabular}{ccccc} 
\hline
\multirow{2}{*}{\textbf{Methods}} & \multicolumn{4}{c}{Average accuracy of all sessions  (\%)}                                                                                                      \\ 
\cline{2-5}
                                  & \textbf{B$0$ $10$ steps}       & \textbf{B$0$ $20$ steps}                        & \textbf{B$50$ $10$ steps}       &  \textbf{B$50$ $25$ steps}         \\ 
\hline
Bound                             & $80.40$                         & $80.41$                                          & $81.49$                          & $81.74$                           \\ 
\hline
iCaRL~\cite{icarl}    & $64.42$  & $63.5$  & $53.78$ & $50.60$\\
BiC~\cite{bic}      & $65.08$  & $62.37$  & $53.21$  &  $48.96$\\
WA~\cite{WA}       & $67.08$  & $64.64$  & $57.57$  & $54.10$\\
COIL\cite{zhou2021co} & $65.48$ & $62.98$ & $59.96$ & - \\
PODNet~\cite{douillard2020podnet}   & $55.22$  & $47.87$  &  $63.19$  & $60.72$\\
DER~\cite{der}     & $69.74$ & $67.98$  & $66.36$  & -\\
\hline
Ours      & $\textbf{72.90}$& $\textbf{70.65}$& $\textbf{67.95}$ &  $\textbf{63.83}$\\ 
Improvement            & $\gain{3.06}$ & $\gain{2.67}$ & $\gain{1.59}$ & $\gain{3.11}$ \\ \hline
\end{tabular}
\end{table*}
\noindent\textbf{Implementation Details.}
Our method and all compared methods are implemented with Pytorch~\cite{pytorch} and PyCIL~\cite{zhou2021pycil}. For ImageNet, we adopt the standard ResNet-18~\cite{resnet} as our feature extractor and set the batch size to 256. The learning rate starts from 0.1 and gradually decays to zero with a cosine annealing scheduler~\cite{loshchilov2016sgdr}~(170 epochs in total). For CIFAR-100, we use a modified ResNet-32~\cite{icarl} as the most previous works as our feature extractor and set the batch size to 128. The learning rate also starts from 0.1 and gradually decays to zero with a cosine annealing scheduler~(170 epochs in total). For both ImageNet and CIFAR-100, we use SGD with the momentum of 0.9 and the weight decay of 5e-4 in the boosting stage. In the compression stage, we use SGD with the momentum of 0.9 and set the weight decay to 0. We set the temperature scalar $T$ to 2. For data augmentation, AutoAugment~\cite{cubuk2019autoaugment}, random cropping, horizontal flip, and normalization are employed to augment training images. The hyperparameter $\beta$ in Eq.~\ref{eq:bkd} is set to 0.97 in most settings, while the $\beta$ in Eq.~\ref{eq:la} on CIFAR-100 and ImageNet-100/1000 is set to 0.95 and 0.97, respectively. 

\subsection{Quantitative results}
\noindent\textbf{CIFAR-100.}
Table~\ref{tab:cifar100} and Fig.~\ref{fig:cifar100} summarize the results of CIFAR-100 benchmark. We use replay as the baseline method, which only uses rehearsal strategy to alleviate forgetting. Experimental results show that our method outperforms the other state-of-the-art strategies in all six settings on CIFAR-100. Our method achieves excellent performance on both long-term incremental learning tasks and large-step incremental learning tasks. Particularly, we achieve 3.11\% and 2.67\% improvement under the long-term incremental setting of base 50 with 25 steps and base 0 with 20 steps, respectively. We also surpass the state-of-the-art method by 1.71\% and 3.06\% under the large step incremental learning setting of 20 classes per step and 10 classes per step.  It should also be noted that although our method FOSTER expands a new module every time, we compress it to a single backbone every time. Therefore, the parameters and feature dimensions of our model do not increase with the number of tasks, which is our advantage over methods~\cite{der,simpleder,dytox} based on dynamic architecture. From Fig.~\ref{fig:cifar100}, we can see that the compressed single backbone model FOSTER has a tiny gap with FOSTER B4 in each step, which verifies the effectiveness of our distillation method.

\begin{figure*}[t]
  \centering
  \begin{subfigure}[b]{0.32\linewidth}
    \includegraphics[width=\linewidth]{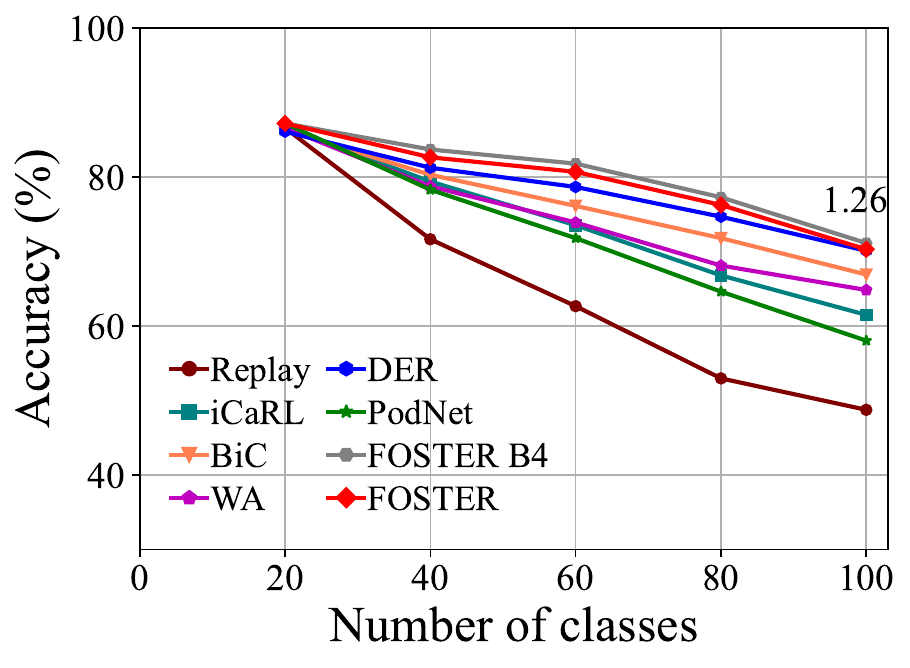}
    \caption{\scriptsize ImageNet-100  B0   5  steps}
    \label{fig:imagenet_inc1}
  \end{subfigure}
  \hfill
  \begin{subfigure}[b]{0.32\linewidth}
    \includegraphics[width=\linewidth]{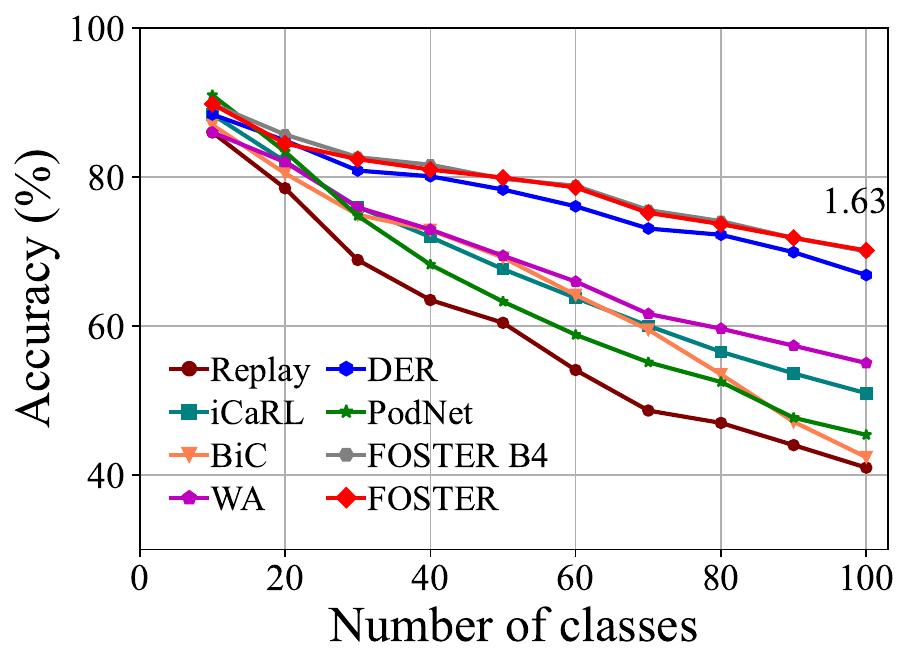}
    \caption{\scriptsize ImageNet-100  B0   10  steps}
    \label{fig:imagenet_inc2}
  \end{subfigure}
   \hfill
  \begin{subfigure}[b]{0.32\linewidth}
    \includegraphics[width=\linewidth]{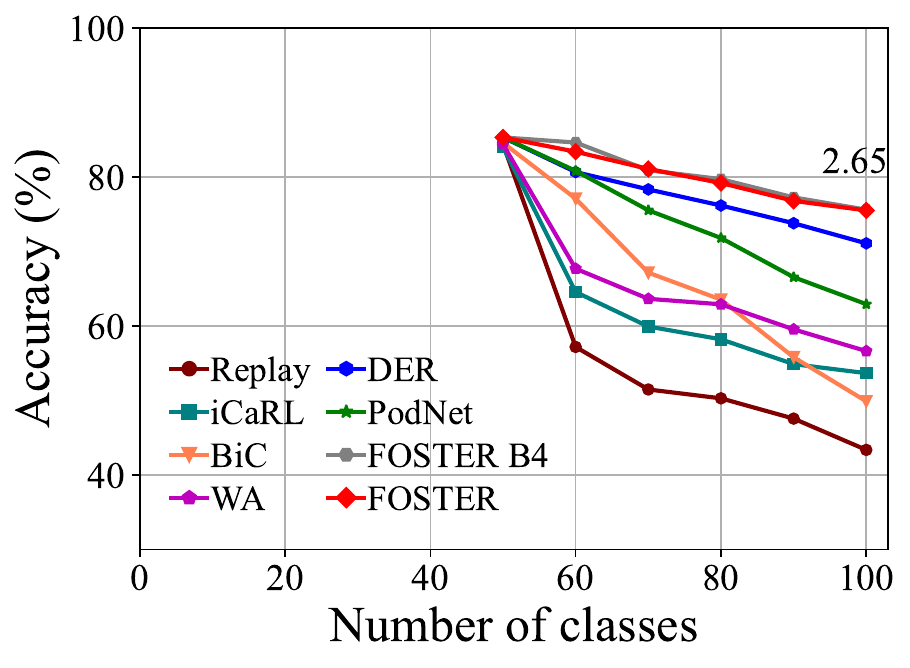}
    \caption{\scriptsize ImageNet-100  B50  5  steps}
    \label{fig:imagenet_inc3}
  \end{subfigure}
  \caption{\small  \textbf{Incremental Accuracy on ImageNet-100.} Replay is the baseline with naive rehearsal strategy. FOSTER B4 records the accuracy of the dual branch model after feature boosting. FOSTER records the accuracy of the single backbone model after feature compression. The performance gap is annotated at the end of each curve.}
  \label{fig:ImageNet}
\end{figure*}

\begin{table}[t]
\centering
\caption{\small  Average incremental accuracy on ImageNet for FOSTER vs. state-of-the-art. DER uses the same number of backbone models as incremental sessions, while the other methods, including FOSTER, retain only one backbone after each session. The left three columns are experimental results on ImageNet-100. The rightmost column is the results of ImageNet-1000 with 100 classes per step~(10 steps in total).}
\label{tab:ImageNet}

\setlength{\tabcolsep}{2mm}
\begin{tabular}{ccccc} 
\hline
\multirow{2}{*}{\textbf{Methods}} & \multicolumn{4}{c}{Average accuracy of all sessions  (\%)}                                                                                                      \\ 
\cline{2-5}
                                  & \textbf{B$0$ $20$ steps}        & \textbf{B$50$ $10$ steps}                        & \textbf{B$50$ $25$ steps}        & \textbf{ImageNet-1000}            \\ 
\hline
Bound                             & $81.20$                         & $81.20$                                          & $81.20$                          & $89.27$                           \\ 
\hline
iCaRL~\cite{icarl}                           & $62.36$                         & $59.53$                                          & $54.56$                          & $38.4$                            \\
BiC~\cite{bic}                              & $58.93$                         & $65.14$                                          & $59.65$                          & -                                 \\
WA~\cite{WA}                              & $63.2$                          & $63.71$                                          & $ 58.34$                         & $54.10$                           \\
PODNet~\cite{douillard2020podnet}                           & $53.69$                         & $74.33$                                          & $67.28$                          & -                                 \\
DER~\cite{der}                              & $73.79$                         & $77.73$                                          & -                                & $66.73$                           \\ 
\hline
Ours                              & $\textbf{74.49}$                & $\textbf{77.54}$                                 & $\textbf{69.34}$                 & $\textbf{68.34}$                  \\
Improvement            & $\gain{0.7}$ & $\textcolor{gray}{(-0.19)}$ & $\gain{2.06}$ & $\gain{1.61}$ \\\hline
\end{tabular}
\end{table}

\noindent\textbf{ImageNet.}
Table~\ref{tab:ImageNet} and Fig.~\ref{fig:ImageNet} summarize the experimental results for ImageNet-100 and ImageNet-1000 benchmarks. Our method, FOSTER, still outperforms the other method in most settings. In the setting of ImageNet-100 B0, we surpass the state-of-the-art method by 1.26, 1.63, and 0.7 percent points for, respectively, 5, 10, and 20 steps. The results shown in Fig.~\ref{fig:ImageNet} again verify the effectiveness of our distillation strategy, where the performance degradation after compression is negligible.  The results on ImageNet-1000 benchmark is shown in the rightmost column in Tabel~\ref{tab:ImageNet}. Our method improves the average top-1 accuracy on ImageNet-1000 with 10 steps from 66.73\% to 68.34\%~(+1.61\%), showing that our method is also efficacious in large-scale incremental learning.

\subsection{Ablation Study}

\begin{figure}[t]
    \centering
  \begin{subfigure}[b]{0.32\linewidth}
    \includegraphics[width=\linewidth]{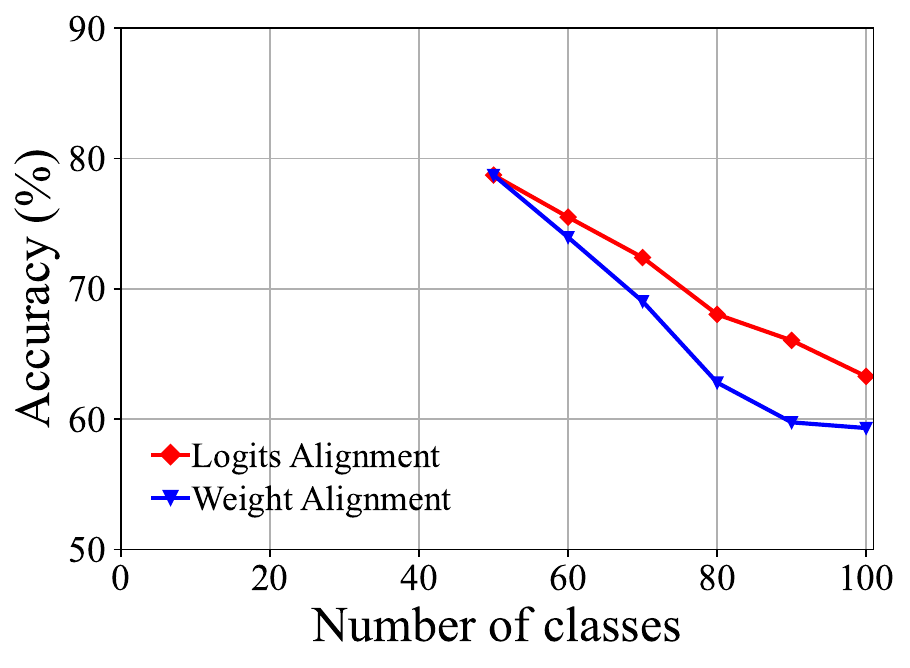}
    \caption{\small Logits alignment}
    \label{fig:LA}
  \end{subfigure}
  \hfill
  \begin{subfigure}[b]{0.32\linewidth}
    \includegraphics[width=\linewidth]{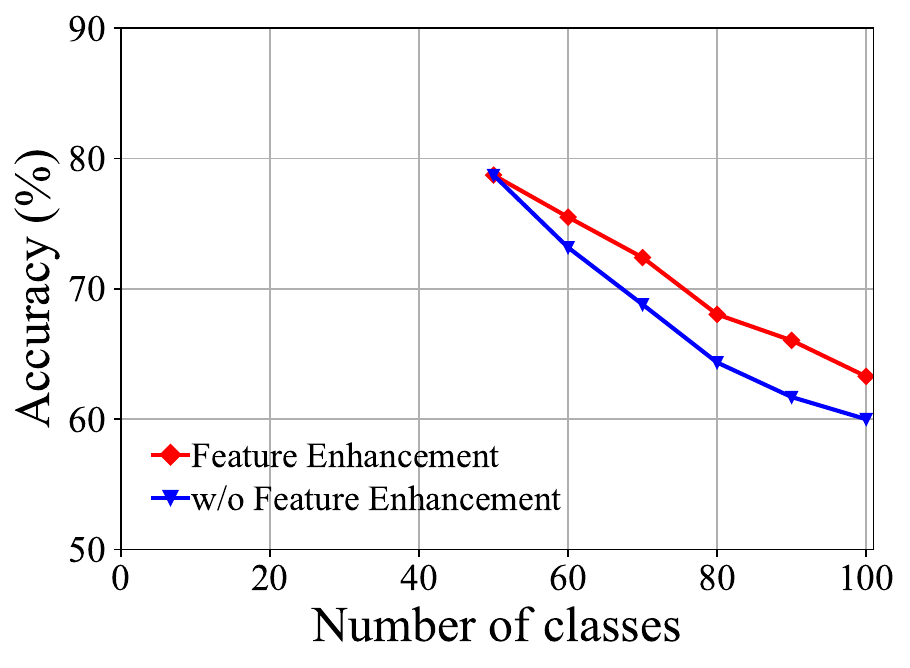}
    \caption{\small Feature enhancement}
    \label{fig:FE}
  \end{subfigure}
   \hfill
  \begin{subfigure}[b]{0.32\linewidth}
    \includegraphics[width=\linewidth]{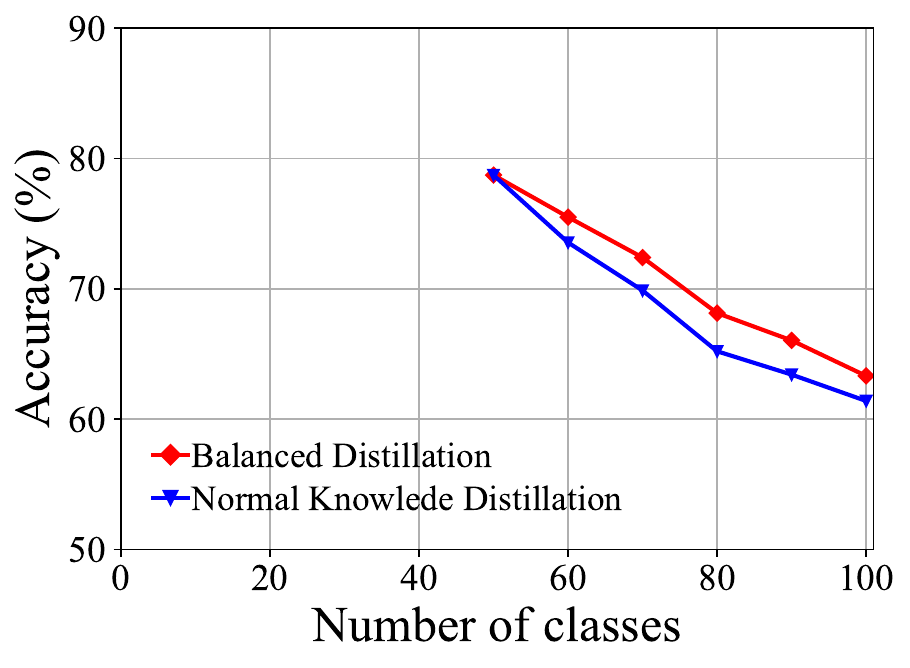}
    \caption{\small Balanced distillation}
    \label{fig:BKD}
  \end{subfigure}
    \caption{\small \textbf{Ablations} of the different key components of FOSTER. (a): Performance comparison between logits alignment and weight alignment~\cite{WA}. (b): Performance comparison with or without Feature Enhancement. (c): Performance comparison between balanced distillation and normal knowledge distillation~\cite{KD}.} 
    \label{fig:ablation}
\end{figure}

\noindent\textbf{Different Components  of FOSTER.}
Table~\ref{fig:ablation} demonstrates the results of our ablative experiments on CIFAR-100 B50 with 5 steps.  Specifically, we replace logits alignments~(LA) with the post-processing method weight alignment~(WA)~\cite{WA}. The performance comparison is shown in Fig.~\ref{fig:LA}, where LA surpasses WA by about 4\% in the final accuracy. This shows that our LA is a more efficacious strategy than WA in calibration for old and new classes. We remove feature enhancement and compare its performance with the original result in Fig.~\ref{fig:FE}, the model suffers from more than 3\% performance decline in the last stage. We find that, in the last step, there is almost no difference in the accuracy of new classes between the model with feature enhancement and the model without that. Nevertheless, the model with feature enhancement outperforms the model without that by more than 4 \% on old categories, showing that feature enhancement encourages the model to learn more about old categories. We compare the performance of balanced knowledge distillation~(BKD) with that of normal knowledge distillation~(KD) in Fig.~\ref{fig:BKD}. BKD surpasses KD in all stages, showing that BKD is more effective when training on imbalanced datasets.

\begin{figure}[t]
    \centering
    \begin{subfigure}[b]{0.46\linewidth}
    \includegraphics[width=\linewidth]{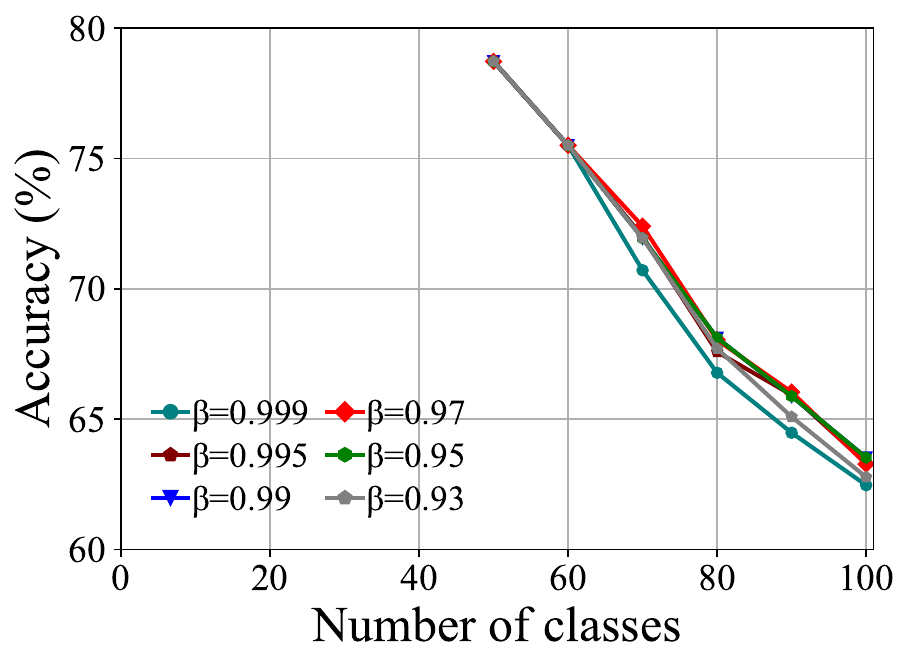}
    \caption{\small Sensitive study of hyper-parameters}
    \label{fig:beta}
  \end{subfigure}
  \hfill
   \begin{subfigure}[b]{0.46\linewidth}
    \includegraphics[width=\linewidth]{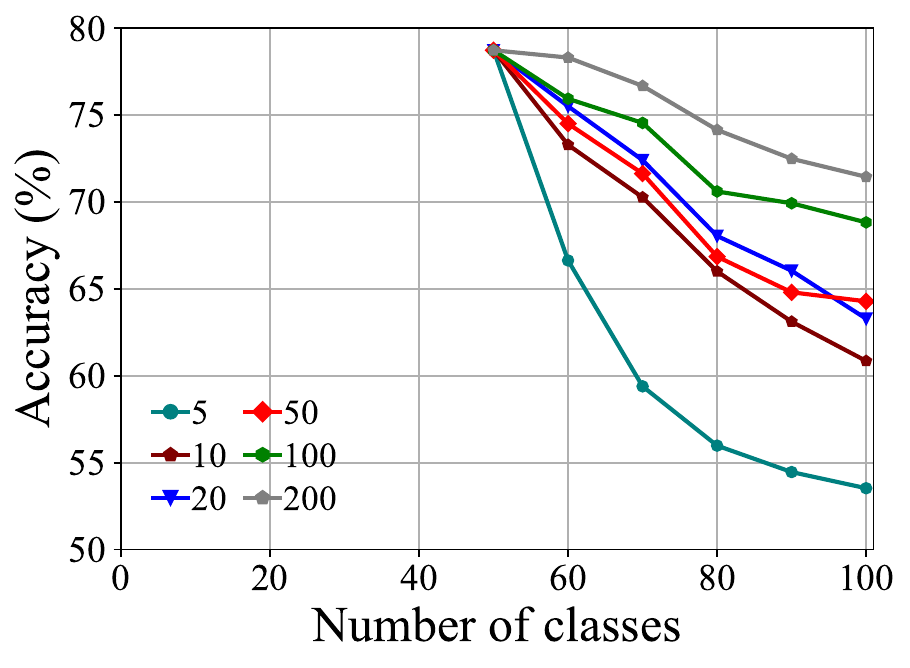}
    \caption{\small Influence of number of exemplars }
    \label{fig:exemplar}
  \end{subfigure}
    \caption{\small  \textbf{Robustness Testing.} Left: Performance under different hyperparameter $\beta$s. Right: Performance with different numbers of exemplars. Both of them are evaluated on CIFAR-100 B50 with 5 steps.}
    \label{fig:RT}
\end{figure}

\noindent\textbf{Sensitive Study of Hyper-parameters.}
To verify the robustness of FOSTER, we conduct experiments on CIFAR-100 B50 5 steps with different hyperparameters $\beta\in (0,1)$. Typically, $\beta$ is set to more than $0.9$. We test $\beta=0.93$, $0.95$, $0.97$, $0.99$, $0.995$, $0.999$ respectively. The experimental results are shown in Fig.~\ref{fig:beta}.  We can see that the performance changes are minimal under different $\beta$s.

\noindent\textbf{Effect of Number of Exemplars.}
In Fig.~\ref{fig:exemplar}, We gradually increase the number of exemplars from 5 to 200 and record the performance of the model on CIFAR-100 B50 with 5 steps. The accuracy in the last step increases from 53.53\% to 71.4\% as the number of exemplars for every class changes from 5 to 200. From the results, we can see that with the increase in the number of exemplars, the accuracy of the last stage of the model gradually improves, indicating that our model can make full use of more exemplars to improve performance. 
In addition, notice that our model achieves more than 60\% accuracy in the last round, even when there are only 10 exemplars for each class, surpassing most state-of-the-art methods using 20 exemplars shown in Fig.~\ref{fig:cifar_inc3}. This indicates that FOSTER is more effective and robust; it can overcome forgetting even with fewer exemplars. 

\begin{figure}[t]
    \centering
    \includegraphics[width=12.2cm]{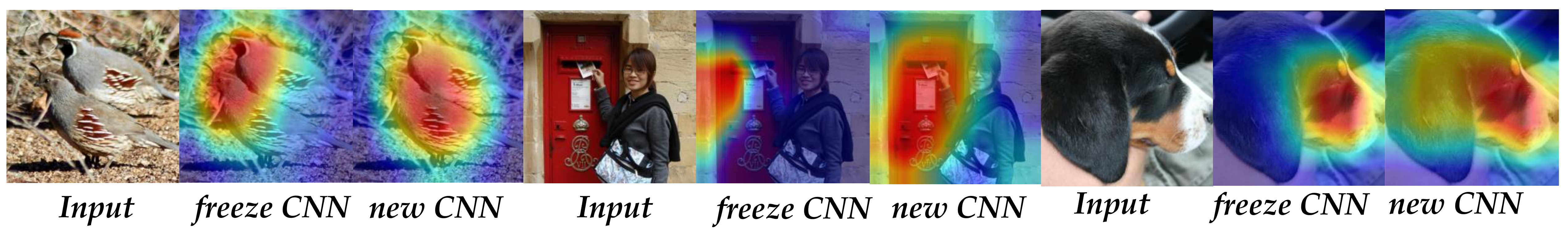}
    \caption{\small \textbf{Grad-CAM before and after feature boosting.} The freeze CNN only focuses on some areas of an object and is not accurate enough, but the new CNN can discover those important but ignored patterns and correct the original output.}
    \label{fig:gradcam}
\end{figure}

\noindent\textbf{Visualization of Grad-CAM.} We visualize the grad-CAM before and after feature boosting. As shown in Fig.~\ref{fig:gradcam}~(left), the freeze CNN only focuses on the head of the birds, ignoring the rest of their bodies, while the new CNN learns that the whole body is important for classification, which is consistent with our claim in Sec.~\ref{sec:intro}. Similarly, the middle and right figures show that the new CNN also discovers some essential but ignored patterns of the mailbox and the dog.

\section{Conclusions}
In this work, we apply the concept of gradient boosting to the scenario of class-incremental learning and propose a novel learning paradigm FOSTER based on that, empowering the model to learn new categories adaptively. At each step, we create a new module to learn residuals between the target and the original model. We also introduce logits alignment to alleviate classification bias and feature enhancement to balance the representation learning of the old and new classes. Furthermore, we propose a simple yet effective distillation strategy to remove redundant parameters and dimensions, compressing the expanded model into a single backbone model. Extensive experiments on three widely used incremental learning benchmarks show that our method obtains state-of-the-art performance. 
\subsubsection*{Acknowledgments.}
This research was supported by National Key
R\&D Program of China (2020AAA0109401), NSFC (61773198, 61921006,62006112), NSFC-NRF Joint Research Project under Grant 61861146001, Collaborative Innovation Center of Novel Software Technology and Industrialization, NSF of Jiangsu Province (BK20200313), CCF-Hikvision Open Fund (20210005). Han-Jia Ye is the corresponding author.

\clearpage
\bibliographystyle{splncs04}
\bibliography{egbib}

\newpage

\setcounter{section}{0}
\renewcommand{\thesection}{\Roman{section}}
\begin{center}
	\textbf{\large Supplementary Material }
\end{center}

\setcounter{equation}{0}
\setcounter{figure}{0}
\setcounter{table}{0}
\setcounter{page}{1}
\section{Rationality Analysis of the Substitution.}\label{sec:1}
We argue that our simplification of replacing the sum of softmax with softmax of logits sum and substituting the distance metric $\textrm{Dis}(\cdot,\cdot)$ for the Kullback-Leibler divergence (KLD) $\textrm{KL}(\cdot\mid\mid\cdot)$. KLD can evaluate the residual between the target and the output by calculating the distance between the target label distribution and the output distribution of categories. KLD is more suitable for classification tasks, and there are some works~\cite{hu2013kullback,rubinstein2004cross} that point out that the KLD has many advantages in many aspects, including faster optimization and better feature representation. Typically, to reflect the relative magnitude of each output, we use non-linear activation softmax to transform the output logits into the output probability. Namely, $p_1,p_2,\dots,p_{\lvert \hat{\mathcal Y}_t\rvert }$, where $0\le p_i\le 1$, $\sum_{i=1}^{\lvert \hat{\mathcal Y}_t}p_{i}=1$ and $\lvert \hat{\mathcal Y}_{t}\rvert$ is the number of all seen categories. In classification tasks, the target label is usually set to 1, and the non-target label is set to 0. Therefore, we expect the output of the boosting model can be constrained between $0$ and $1$.  Simply combining the softmax outputs of the original model $\mathrm F_{t-1}$ and $\mathcal F_{t}$ can not satisfy the constraints. Suppose that the output of $\mathrm F_{t-1}$ and $\mathcal F_t$ in class $i$ are $p_i^o$ and $p_i^n$,  the combination of $p_i^n$ and $p_i^o$ is not in line with our expectation since $0\le p_i^o+p_i^n\le 2$. By replacing the sum of softmax with softmax of logits sum, we can limit the output of the boosting model between 0 and 1, and the judgment of the two models can still be integrated.

\section{Influence of the Initialization of the Weight $\mathbf O$}\label{sec:2}
In this section, we discuss the effect of the initialization of the weight $\mathbf O$ in the super linear classifier of our boosting model.
\begin{align}
    \mathbf W_t^\top&= \left[\begin{array}{cc}\mathbf W_{t-1}^\top & (\mathcal W_{t}^{(o)})^\top\\
\mathbf O& (\mathcal W_{t}^{(n)})^\top\end{array}\right]\,  .
\end{align}

In the main paper, we set $\mathbf O$ to all zero as our default initialization strategy. Therefore, the outputs of the original model for new categories are zero, thus having nothing to do with the classification of new classes. 

Here, we introduce three different initialization strategies, including fine-tune~(FT), all-zero~(AZ), and all-zero with bias~(AZB), to further explore the impact of different initialization strategies on performance. Among them, FT is directly training $\mathbf O$ without any restrictions. AZ sets the outputs of the old model on the new class to all zero, and thus the outputs of the model on the new class logits only contain the output of the new model, and the old model does not provide any judgment on the new class. Based on AZ, AZB adds bias learning to balance the logits of the old and new categories. 
\begin{figure}[t]
    \centering
    \includegraphics[width=0.5\textwidth]{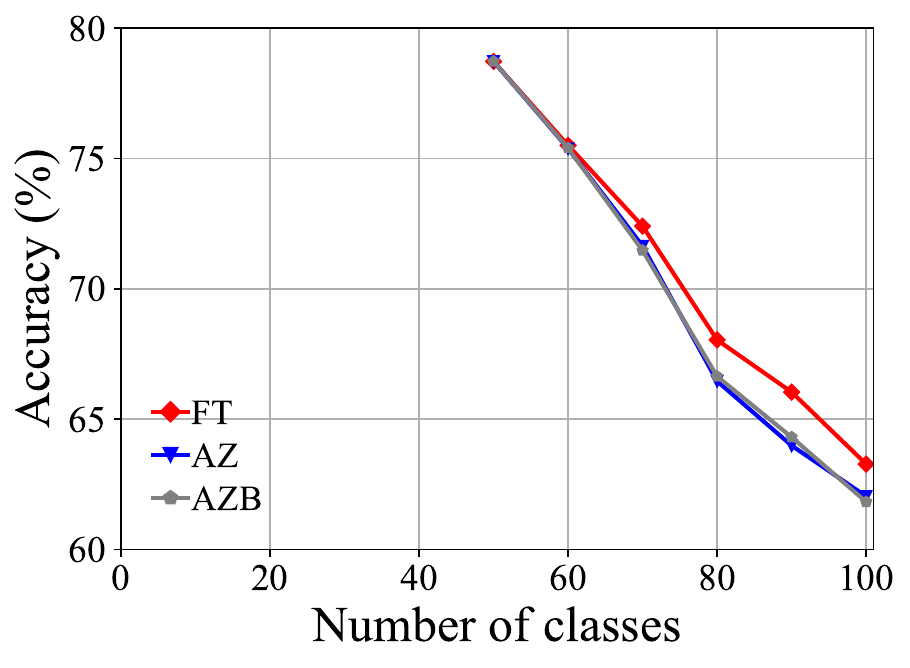}
    \caption{\textbf{Influence of different initialization strategies.}The red line represents FT, the blue line represents AZ, and the gray line represents AZB. 
The performance of FT is slightly better than AZ and AZB. The performance gap between AZ and AZB is negligible.}
    \label{fig:o}
\end{figure}
Fig.~\ref{fig:o} illustrates the comparison of performance on CIFAR-100~\cite{cifar100} B50 with 5 steps with different initialization strategies. We can see that the performance of using FT initialization strategy is slightly better than that of using AZ and AZB initialization strategies, but the difference is not significant. The performance gap between AZ and AZB is negligible, indicating that the influence of bias is weak.

\begin{figure}[t]
    \centering
    \begin{subfigure}[b]{0.31\linewidth}
    \includegraphics[width=\linewidth]{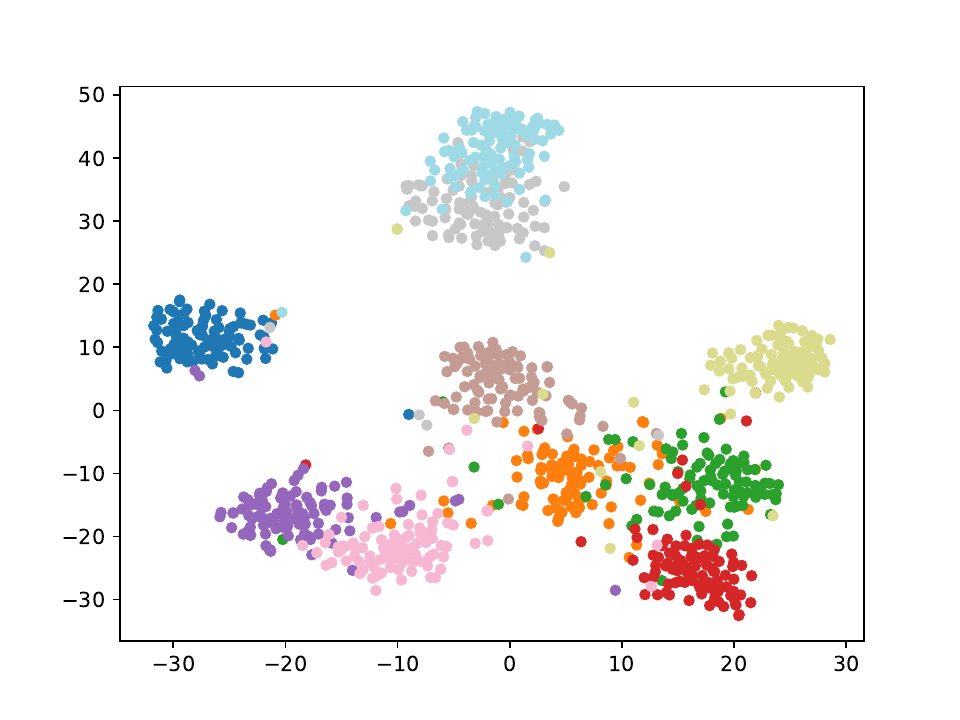}
    \caption{Fine-tune 50 classes}
    \label{fig:finetunef50}
  \end{subfigure}
  \hfill
\begin{subfigure}[b]{0.31\linewidth}
    \includegraphics[width=\linewidth]{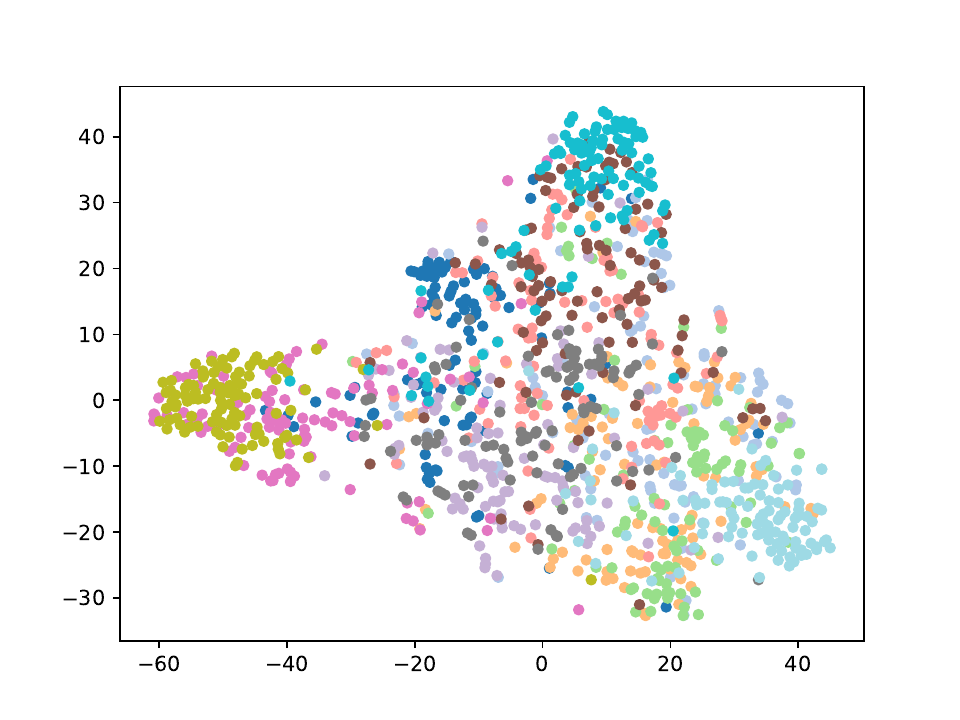}
    \caption{Fine-tune 60 classes }
    \label{fig:finetunef60}
  \end{subfigure}
  \hfill
 \begin{subfigure}[b]{0.31\linewidth}
    \includegraphics[width=\linewidth]{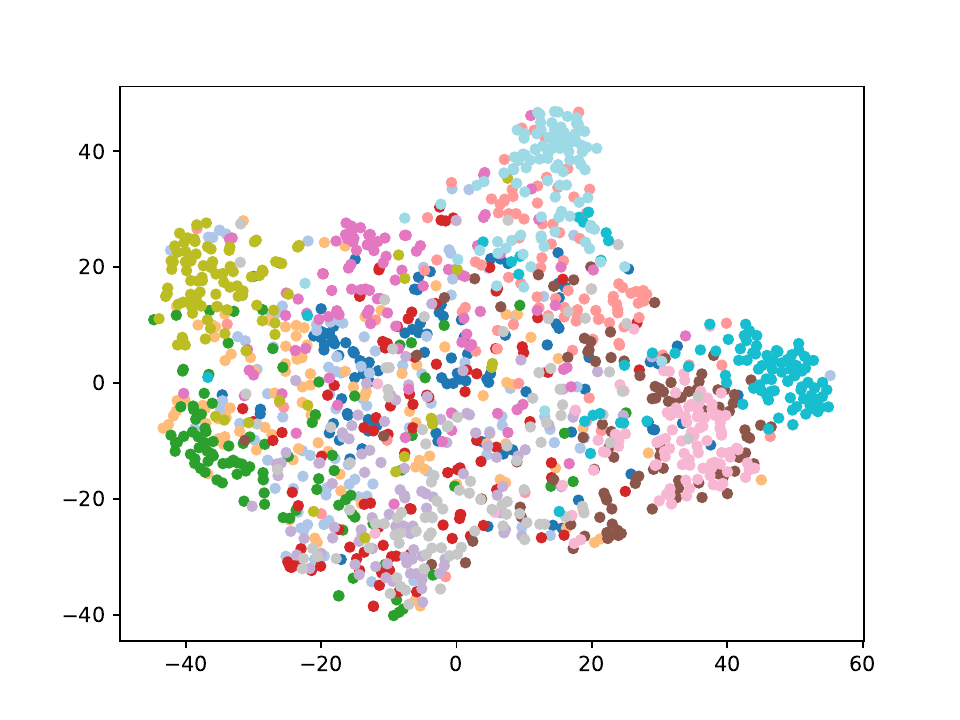}
    \caption{Fine-tune 70 classes }
    \label{fig:finetunef70}
  \end{subfigure}
  \hfill
\begin{subfigure}[b]{0.31\linewidth}
    \includegraphics[width=\linewidth]{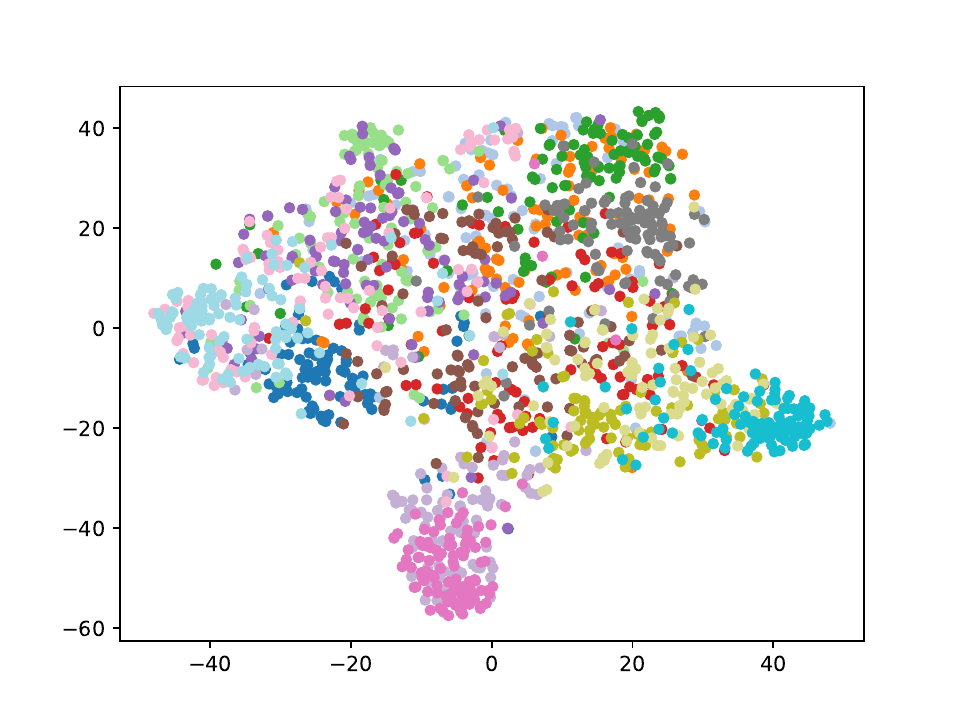}
    \caption{Fine-tune 80 classes }
    \label{fig:finetunef80}
  \end{subfigure}
  \hfill
\begin{subfigure}[b]{0.31\linewidth}
    \includegraphics[width=\linewidth]{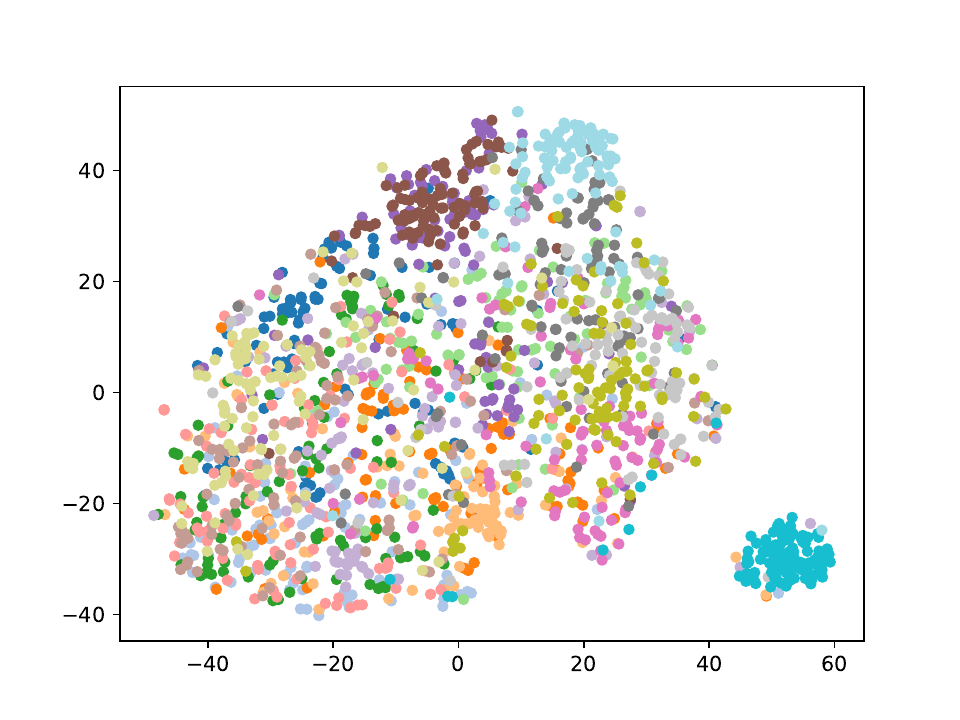}
    \caption{Fine-tune 90 classes }
    \label{fig:finetunef90}
  \end{subfigure}
  \hfill
\begin{subfigure}[b]{0.31\linewidth}
    \includegraphics[width=\linewidth]{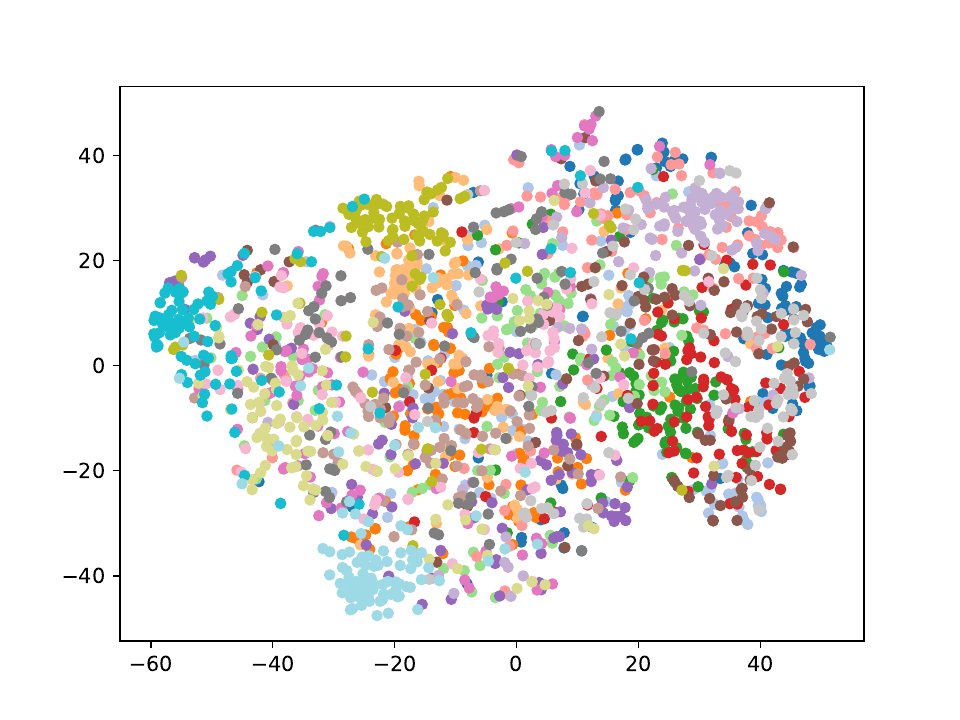}
    \caption{Fine-tune 100 classes }
    \label{fig:finetunef100}
  \end{subfigure}
  \hfill
    \begin{subfigure}[b]{0.31\linewidth}
    \includegraphics[width=\linewidth]{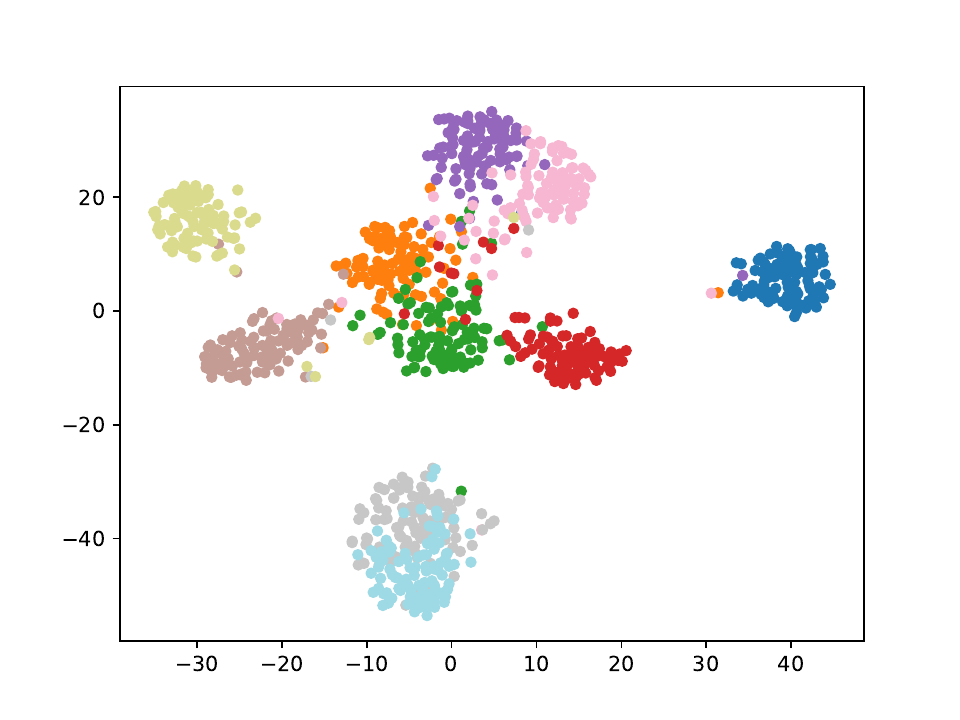}
    \caption{FOSTER 50 classes}
    \label{fig:f50}
  \end{subfigure}
  \hfill
   \begin{subfigure}[b]{0.31\linewidth}
    \includegraphics[width=\linewidth]{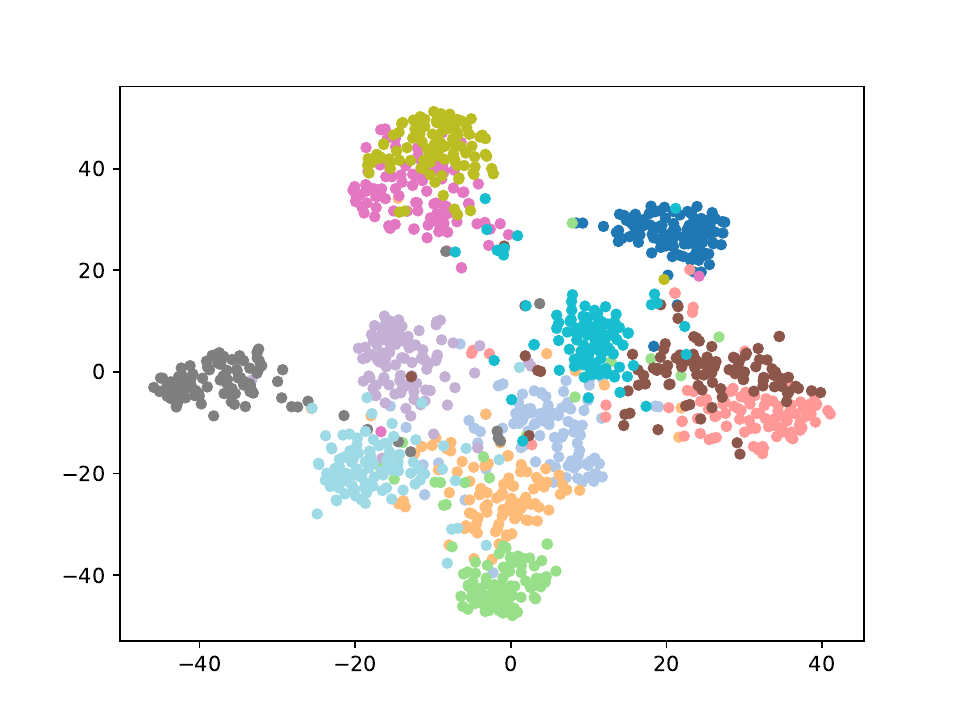}
    \caption{FOSTER 60 classes }
    \label{fig:f60}
  \end{subfigure}
  \hfill
   \begin{subfigure}[b]{0.31\linewidth}
    \includegraphics[width=\linewidth]{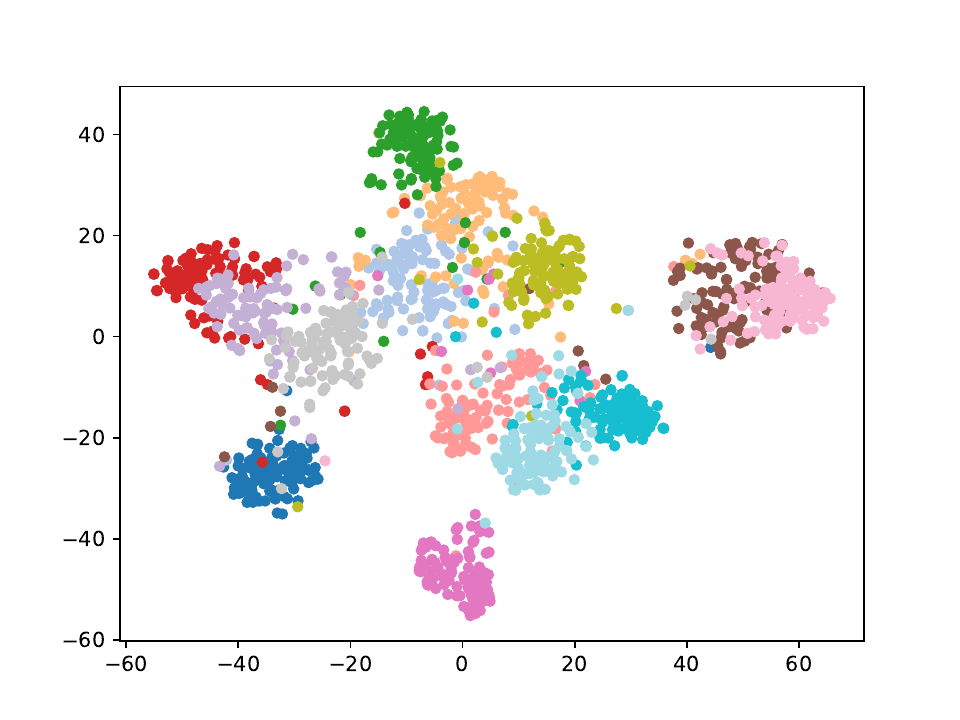}
    \caption{FOSTER 70 classes }
    \label{fig:f70}
  \end{subfigure}

   \begin{subfigure}[b]{0.31\linewidth}
    \includegraphics[width=\linewidth]{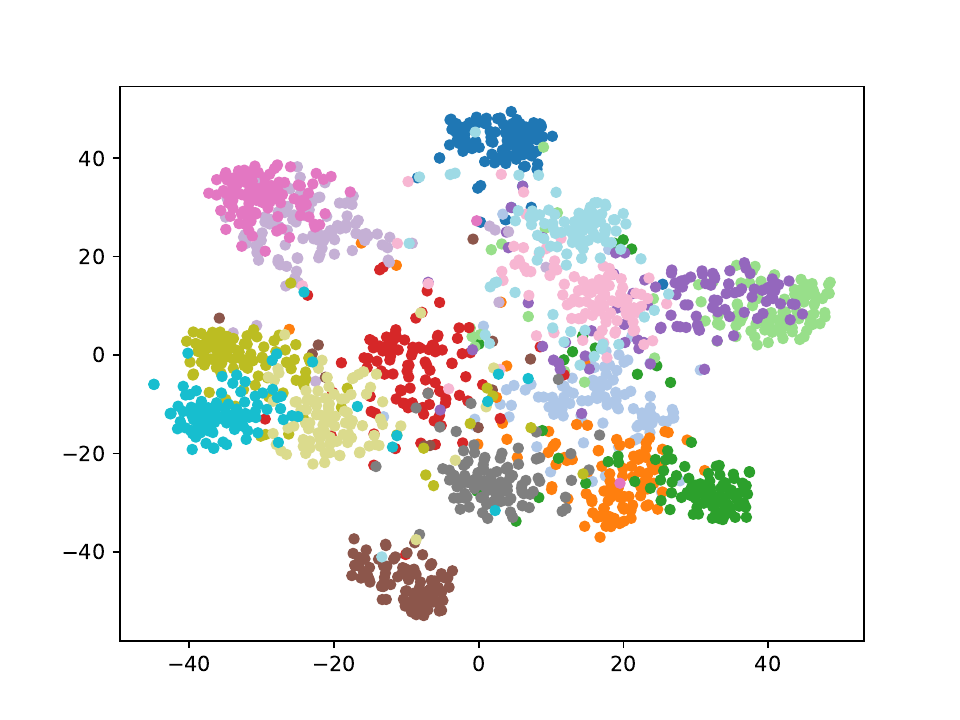}
    \caption{FOSTER 80 classes }
    \label{fig:f80}
  \end{subfigure}
  \hfill
   \begin{subfigure}[b]{0.31\linewidth}
    \includegraphics[width=\linewidth]{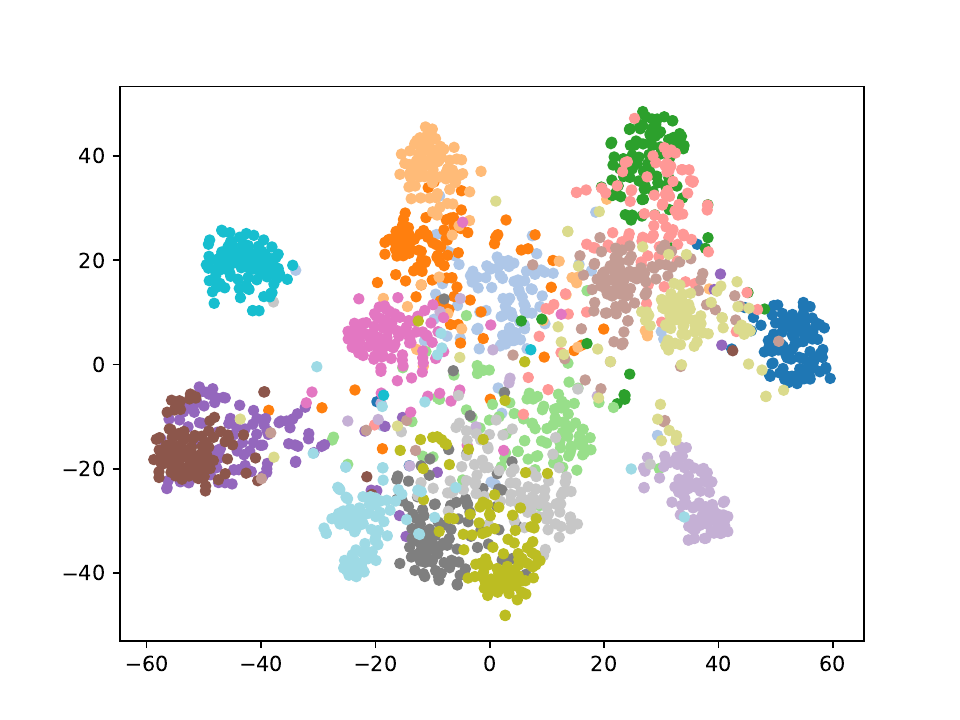}
    \caption{FOSTER 90 classes }
    \label{fig:f90}
  \end{subfigure}
  \hfill
    \begin{subfigure}[b]{0.31\linewidth}
    \includegraphics[width=\linewidth]{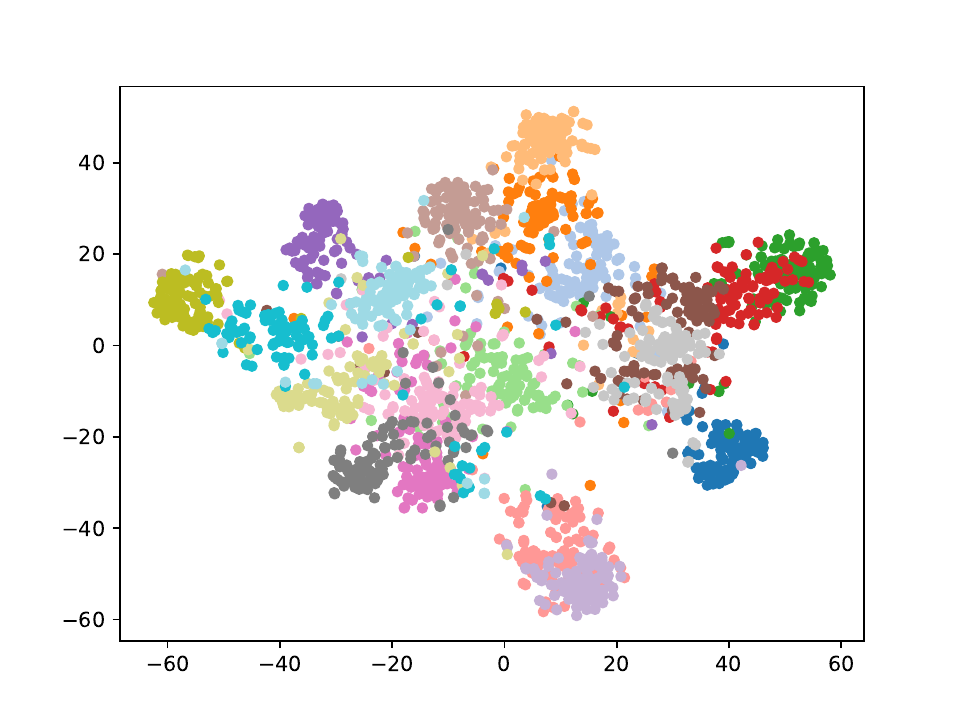}
    \caption{FOSTER 100 classes }
    \label{fig:f100}
  \end{subfigure}
    \caption{\textbf{t-SNE~\cite{van2008visualizing} visualization of CIFAR-100~\cite{cifar100} B50 with 5 steps.} Figure~(a)-(g) shows the t-SNE visualization of fine-tune method. Figure~(h)-(l) shows the t-SNE visualization of our method FOSTER. \textbf{In order to achieve better results, we normalize each feature and randomly select one category in each five categories for visualization.}}
    \label{fig:ft}
\end{figure}

\section{Introduction to Compared Methods}\label{sec:3}
In this section, we will describe in detail the methods compared in the main paper.

\noindent \textbf{Fine-tune:} Fine-tune is the baseline method that simply updates its parameters when a new task comes, suffering from catastrophic forgetting. By default, weights corresponding to the outputs of previous classes in the final linear classifier are not updated.

\noindent \textbf{Replay:} Replay utilizes the rehearsal strategy to alleviate the catastrophic forgetting compared to Fine-tune. We use herding as the default way of choosing exemplars from the old data.

\noindent \textbf{iCaRL~\cite{icarl}:} iCaRL combines cross-entropy loss with knowledge distillation loss together. It retains an old model to help the new model maintain the discrimination ability through knowledge distillation on old categories. To mitigate the classification bias caused by the imbalanced dataset when learning new tasks, iCaRL calculates the center of exemplars for each category and uses NME as the classifier for evaluation.  

\noindent \textbf{BiC~\cite{bic}:} BiC performs an additional bias correction process compared to iCaRL, retaining a small validation set to estimate the classification bias resulting from imbalanced training. The final logits are computed by

\begin{align}
q_{k}=\left\{\begin{array}{lr}
o_{k} & 1 \leq k \leq n \\
\alpha o_{k}+\beta & n+1 \leq k \leq n+m
\end{array}\right. ,
\end{align}
 where $n$ is the number of old categories and $m$ is the number of new ones. the bias correction step is to estimate the appropriate $\alpha$ and $\beta$.

\noindent \textbf{WA~\cite{WA}:} During the process of incremental learning, the norms of the weight vectors of new classes are much larger than those of old classes. Based on that, WA proposes an approach called Weight Alignment to correct the biased weights in the final classifier by aligning the norms of the weight vectors of new classes to those of old classes.

\begin{align}
\widehat{\mathbf{W}}_{n e w}=\gamma \cdot \mathbf{W}_{n e w},
\end{align}
where $\gamma=\frac{\operatorname{Mean}\left(\boldsymbol{N o r m}_{\text {old }}\right)}{\operatorname{Mean}\left(\boldsymbol{N o r m}_{n e w}\right)}.$

\noindent \textbf{PODNet~\cite{douillard2020podnet}:} PODNet proposes a novel spatial-based distillation loss that can be applied throughout the model. PODNet has greater performance on long runs of small incremental tasks.

\noindent \textbf{DER~\cite{der}:} DER preserves old feature extractors to maintain knowledge for old categories. When new tasks come, DER creates a new feature extractor and concatenates it with old feature extractors to form a higher dimensional feature space. In order to reduce the number of parameters, DER uses the pruning method proposed in HAT~\cite{serra2018overcoming}, but the number of parameters still increases with the number of tasks. DER can be seen as a particular case of our Boosting model. When we set the weight $\mathbf O$ of boosting model can be trainable, and remove feature enhancement and logits alignment proposed in the main paper, boosting model can be reduced to DER.

\section{Visualization of Detailed Performance}\label{sec:4}
\noindent \textbf{Visualizing Feature Representation.}
We visualize the feature representations of the test data by t-SNE~\cite{van2008visualizing}. Fig.~\ref{fig:ft} illustrates  the comparison of baseline method, fine-tune, with our FOSTER in the setting of CIFAR-100~\cite{cifar100} B50 with 5 steps. As shown in Fig.~\ref{fig:finetunef50} and Fig.~\ref{fig:f50}, in the base task, all categories can form good clusters with explicit classification boundaries. However, as shown in Fig.~\ref{fig:finetunef60}, Fig.~\ref{fig:finetunef70}, Fig.~\ref{fig:finetunef80}, Fig.~\ref{fig:finetunef90}, and Fig.~\ref{fig:finetunef100}, in stages of incremental learning, the result of category clustering becomes very poor without clear classification boundaries. In the last stage which is shown in Fig.\ref{fig:finetunef100}, feature points of each category are scattered. On the contrary, as shown in Fig.~\ref{fig:f50}, Fig.~\ref{fig:f60}, Fig.~\ref{fig:f70}, Fig.~\ref{fig:f80}, Fig.~\ref{fig:f90}, and Fig.~\ref{fig:f100}. our FOSTER method can make all categories form good clusters at each incremental learning stage, and has a clear classification boundary, indicating that our FOSTER method is a very effective strategy in feature representation learning and overcoming catastrophic forgetting.

\begin{figure}[t]
    \centering
    \begin{subfigure}[b]{0.48\linewidth}
    \includegraphics[width=\linewidth]{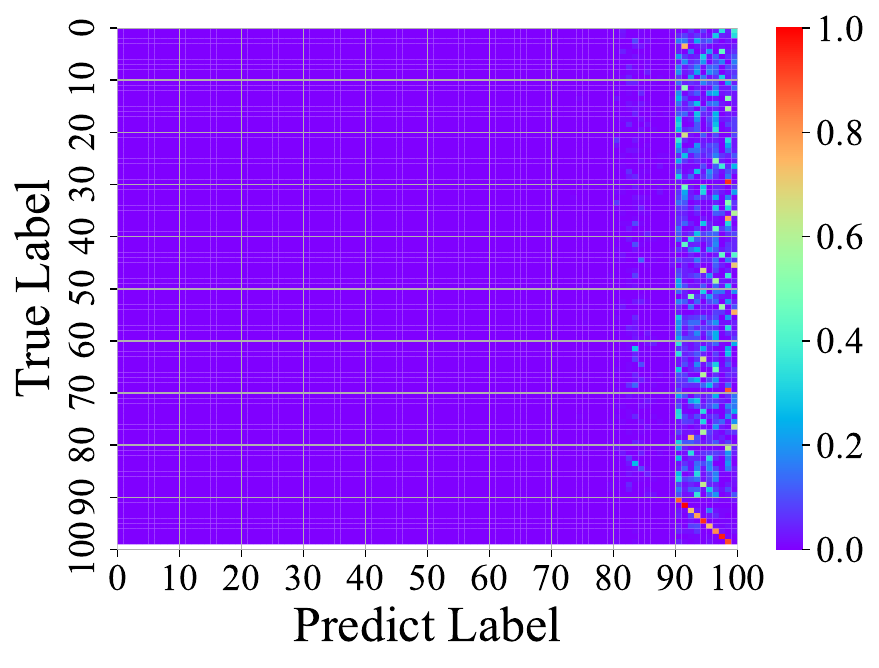}
    \caption{Fine-tune}
    \label{fig:FinetuneCM}
  \end{subfigure}
  \hfill
   \begin{subfigure}[b]{0.48\linewidth}
    \includegraphics[width=\linewidth]{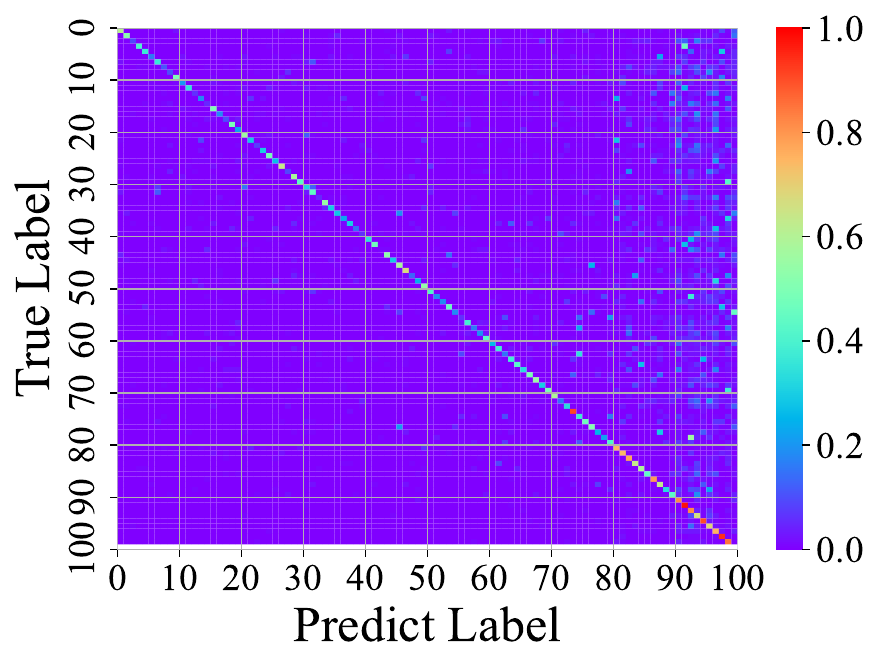}
    \caption{iCaRL}
    \label{fig:iCaRLCM}
  \end{subfigure}
  \hfill
   \begin{subfigure}[b]{0.48\linewidth}
    \includegraphics[width=\linewidth]{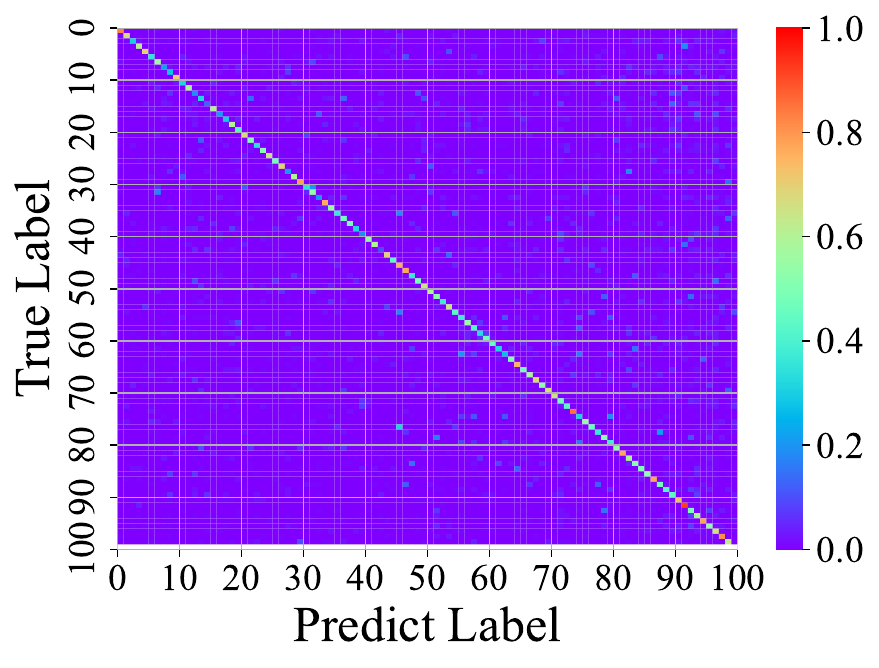}
    \caption{WA}
    \label{fig:WACM}
  \end{subfigure}
  \hfill
   \begin{subfigure}[b]{0.48\linewidth}
    \includegraphics[width=\linewidth]{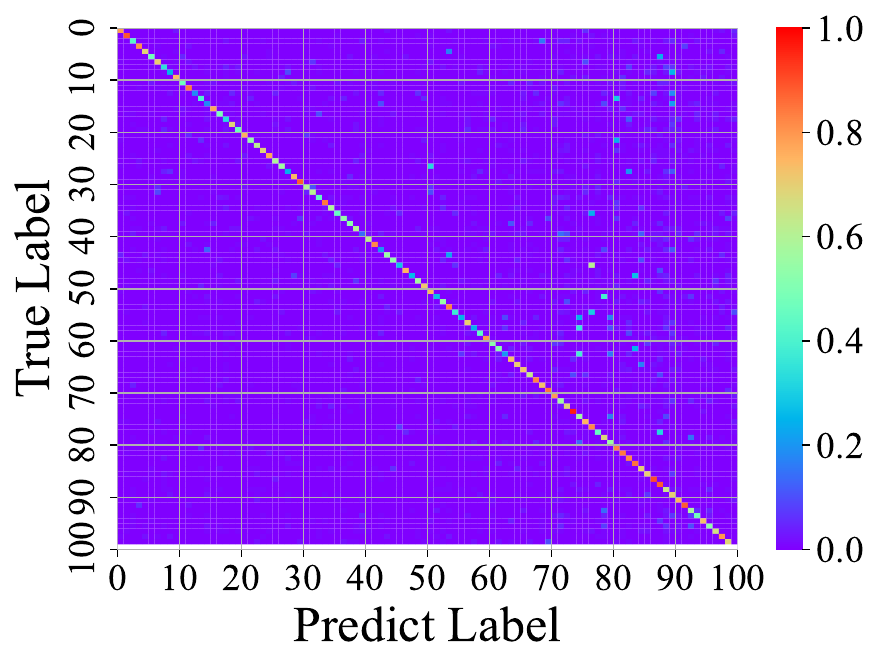}
    \caption{DER}
    \label{fig:DERCM}
  \end{subfigure}
  \hfill
   \begin{subfigure}[b]{0.48\linewidth}
    \includegraphics[width=\linewidth]{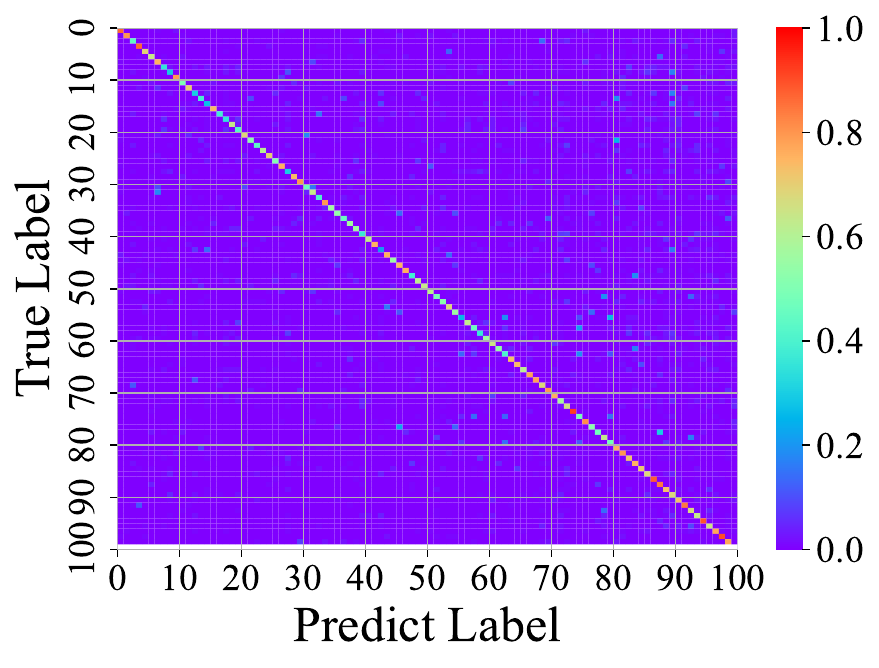}
    \caption{FOSTER}
    \label{fig:FOSTERCM}
  \end{subfigure}
    \caption{ \textbf{Confusion matrices of different methods.} The  vertical  axis   represents  the  real  label,  and  the  horizontal  axis represents the label predicted by the model. The warmer the color of a point in the graph, the more samples it represents.}
    \label{fig:compareCM}
\end{figure}

\noindent \textbf{Visualizing Confusion Matrix.}
To compare with other methods, we visualize the confusion matrices of different methods at the last stage in Fig.~\ref{fig:compareCM}. In these confusion matrices, the vertical axis represents the real label, and the horizontal axis represents the label predicted by the model. Warmer colors indicate higher prediction rates, and cold colors indicate lower ones. Therefore, the warmer the point color on the diagonal and the colder the color on the other points, the better the performance of the model. Fig.~\ref{fig:FinetuneCM} shows the confusion matrix of fine-tune. The brightest colors on the right and colder colors elsewhere suggest that the fine-tune method has a strong classification bias, tending to classify inputs into new categories and suffering from severe catastrophic forgetting. Fig.~\ref{fig:iCaRLCM} shows the confusion matrix of iCaRL~\cite{icarl}. iCaRL has obvious performance improvement compared with fine-tune. However, the columns on the right are still bright, indicating that they also have a strong classification bias. In addition, the points on the diagonal have obvious discontinuities, indicating that they cannot make all categories achieve good accuracy. Fig.~\ref{fig:WACM} shows the confusion matrices of WA~\cite{WA}.  Benefiting from Weight Alignment, WA significantly reduces classification bias compared with iCaRL. The rightmost columns have no obvious brightness. Nevertheless, its accuracy in old classes is not high enough. As shown in the figure, most of his color brightness at the diagonal position of the old class is between 0.2 and 0.4. Fig.~\ref{fig:DERCM} shows the confusion matrices of DER~\cite{der}. DER achieves good results in both old and new categories, but the brightness of the upper right corner shows that it still suffers from classification bias and has room for improvement. As shown in Fig.~\ref{fig:FOSTERCM}, our method FOSTER performs well in all categories and well balances the accuracy of the old and new classes.

\end{document}